\documentclass[preprint,12pt]{elsarticle}
\usepackage[top=1in,bottom=1in,left=1in,right=1in]{geometry}
\usepackage[utf8]{inputenc} % allow utf-8 input
\usepackage[T1]{fontenc}    % use 8-bit T1 fonts
\usepackage{hyperref}       % hyperlinks
\usepackage{url}            % simple URL typesetting
\usepackage{booktabs}       % professional-quality tables
\usepackage{amsfonts}       % blackboard math symbols
\usepackage{nicefrac}       % compact symbols for 1/2, etc.
\usepackage{microtype}      % microtypography
\usepackage{lipsum}
\usepackage{soul,xcolor}
\usepackage{blkarray}

\usepackage{caption}
\usepackage{subcaption}
\usepackage{float}
\usepackage{amsmath}
\usepackage[noabbrev]{cleveref}
\creflabelformat{equation}{#2\textup{#1}#3}
\usepackage{wrapfig}
\usepackage{mathtools}
\usepackage{cancel}
\usepackage{fancyhdr}       % header
\usepackage{graphicx} 
\usepackage[utf8]{inputenc}
\usepackage{lineno}
\usepackage{cleveref}
\usepackage{multirow}

\usepackage[normalem]{ulem}
\usepackage{xcolor}

\biboptions{sort&compress}

\usepackage{amsthm}
\theoremstyle{remark}

\begin{document}

\begin{frontmatter}
  
%% Title
\title{Fast meta-solvers for 3D complex-shape scatterers using neural operators trained on a non-scattering problem}

\author[brown]{Youngkyu Lee} 
\author[brown]{Shanqing Liu}
\author[brown]{Zongren Zou}
\author[brown]{Adar Kahana}
\author[tel]{Eli Turkel}
\author[ansys]{Rishikesh Ranade}
\author[ansys]{Jay Pathak}
\author[brown]{George Em Karniadakis \fnref{2}}

%\fntext[1]{These three authors contributed equally.}
\fntext[2]{Corresponding author: george\_karniadakis@brown.edu (George Em Karniadakis).}
\address[brown]{Division of Applied Mathematics, Brown University, Providence, RI 02906, USA}
\address[tel]{Department of Applied Mathematics, Tel Aviv University, Tel Aviv 69978, Israel}
\address[ansys]{CTO Office, Ansys Inc, Canonsburg, PA 15317, USA}
%\address[eng]{School of Engineering, Brown University, Providence, RI 02906, USA}

% \author{
%   Zongren Zou, Adar Kahana, Enrui Zhang, George Em Karniadakis \\
%   Division of Applied Mathematics \\
%   Brown University \\
%   Providence, RI \\
%   \texttt{\{zongren\_zou, adar\_kahana, enrui\_zhang,  george\_karniadakis\}@brown.edu}  \\
%   \And
%   Eli Turkel \\
%   Department of Applied Mathematics \\
%   Tel Aviv University \\
%   Tel Aviv \\
%   \texttt{eliturkel@gmail.com } \\
%   \And
%   Rishikesh Ranade,  Jay Pathak \\
%   CTO Office \\
%   Ansys Inc \\
%   Canonsburg, PA \\
%   \texttt{\{rishikesh.ranade, jay.pathak\}@ansys.com}
% }

% \begin{document}
% \maketitle

\begin{abstract}
Three-dimensional target identification using scattering techniques requires high accuracy solutions and very fast computations for real-time predictions in some critical applications. 
We first train a deep neural operator~(DeepONet) to solve wave propagation problems described by the Helmholtz equation in a domain \textit{without scatterers} but at different wavenumbers and with a complex absorbing boundary condition. 
% We first train a deep neural operator~(DeepONet) as a \textit{foundation model} to solve wave propagation problems described by the Helmholtz equation in a domain \textit{without scatterers} but at different wavenumbers and with a complex absorbing boundary condition. 
We then design two classes of fast meta-solvers by combining DeepONet with either relaxation methods, such as Jacobi and Gauss-Seidel, or with Krylov methods, such as GMRES and BiCGStab, using the trunk basis of DeepONet as a coarse-scale preconditioner. 
We leverage the spectral bias of neural networks to account for the lower part of the spectrum in the error distribution while the upper part is handled inexpensively using relaxation methods or fine-scale preconditioners. 
The meta-solvers are then applied to solve scattering problems with different shape of scatterers, at no extra training cost. 
We first demonstrate that the resulting meta-solvers are shape-agnostic, fast, and robust, whereas the standard standalone solvers may even fail to converge without the DeepONet. We then apply both classes of meta-solvers to scattering from a submarine, a complex three-dimensional problem. We achieve very fast solutions, especially with the DeepONet-Krylov methods, which require orders of magnitude fewer iterations than any of the standalone solvers. 

%We extend a recently proposed machine-learning-based iterative solver, i.e. the hybrid iterative transferable solver (HINTS), to solve the scattering problem described by the Helmholtz equation in an exterior domain with a complex absorbing boundary condition. The HINTS method combines neural operators (NOs) with standard iterative solvers, e.g. Jacobi and Gauss-Seidel (GS), to achieve better performance by leveraging the spectral bias of neural networks. In HINTS, some iterations of the conventional iterative method are replaced by inferences of the pre-trained NO. In this work, we employ HINTS to solve the scattering problem for both 2D and 3D problems, where the standard iterative solver fails. We consider square and triangular scatterers of various sizes in 2D, and a cube and a model submarine in 3D. We explore and illustrate the extrapolation capability of HINTS in handling diverse geometries of the scatterer, which is achieved by training the NO on non-scattering scenarios and then deploying it in HINTS to solve scattering problems. The accurate results demonstrate that the NO in HINTS method remains effective without retraining or fine-tuning it whenever a new scatterer is given. Taken together, our results highlight the adaptability and versatility of the extended HINTS methodology in addressing diverse scattering problems.
\end{abstract}


\end{frontmatter}

\newpage
% \linenumbers
\section{Introduction}

\subsection{Motivation and background}

Solving computationally partial differential equations (PDEs) that arise in scientific problems forms a cornerstone, bridging applied mathematical modeling with physical and engineering applications. The need for effective numerical solvers for linear and nonlinear systems, following the discretization of equations, has led to many
advances in effective solvers over the last few decades.
Historically, simple discretized systems have been solved using fixed-point iterative solvers, such as Jacobi and Gauss-Seidel methods( see, for instance,~\cite{saad2003iterative}) and their variants Successive Over-Relaxation (SOR) and Symmetric SOR (SSOR) (see, for instance,  \cite{hadjidimos2000successive}) methods. For large-scale systems, Krylov methods, such as Conjugate gradient (CG), Generalized Minimal Residual (GMRES) and Biconjugate gradient stabilized (BiCGStab) methods are now more widely used (see, for instance,~\cite{saad1986gmres,van1992bi}).

There has been a paradigm shift in solving computationally scientific problems in recent years.
The introduction of scientific machine learning~(SciML) has unlocked a large variety of new methods for solving PDEs, including linear and non-linear systems, forward and inverse problems \cite{karniadakis2021physics, cuomo2022scientific, toscano2024pinns}.
Most of these methods replace the traditional solver with ones based on modern machine learning~(ML) techniques. 
One of the most popular SciML approaches proposed recently to tackle PDEs is the physics-informed neural network~(PINN)~\cite{raissi2019physics}, which embeds the governing equations (physics, chemistry, biology, etc.)~\cite{cai2021fluid, cai2021heat, mao2020physics, shukla2025neurosem, pang2019fpinns,lee2024two, linka2022bayesian, zhang2024discovering, mishra2021physics, zou2023hydra, misyris2020physics, zou2024correcting, yin2023generative, sahli2020physics, chen2024leveraging, chen2023leveraging, henkes2022physics, zou2025learning, chiu2022can, zou2024leveraging} into the loss function of a neural network~(NN). This enables the network to incorporate the physics during training and capture the solution once the training converges. The recent use of scaled second-order optimization methods \cite{kiyani2025optimizer} has increased greatly
the accuracy of PINNs.
%PINN is an example of the replacement of conventional numerical solvers since they obtain the sought solution of a PDE by training NNs without invoking any classical solver. 
Another category is operator learning~\cite{lu2021learning, li2021fourier, wang2021learning, kovachki2023neural, lu2022comprehensive, ovadia2023vito, he2023novel, zou2023uncertainty, li2023phase, psaros2023uncertainty, benitez2024out, li2023fourier, zhu2023reliable, zou2024neuraluq, kovachki2021universal, wen2022u, lu2022multifidelity}, where NNs are used to approximate the solution operator of a family of solutions of a PDE via learning the mapping between functions.
Examples include mapping material properties to a PDE solution, a future state to an initial condition~(solving an inverse problem), and one state variable to another.
One of the major advantages of such neural operators~(NOs) is that once trained, the network can infer solutions in real time without retraining.

The effectiveness of SciML methods varied across different problems, especially those with multiscale characteristics. 
%While performing well on various tasks, by completely replacing traditional solvers, centuries of studies and developments in numerical analysis are simply ignored. 
On the other hand, traditional solvers offer some special attributes and high-accuracy solutions, particularly for problems with high-frequency content, at a relatively low cost.
Hence, it may be beneficial to explore \textit{hybrid} methods that combine classical techniques with ML to create uniformly powerful solvers that leverage the strengths of both fields. 
Motivated by this, 
a recent development is the hybrid preconditioning strategy, that uses the pre-trained neural operator as the preconditioner of the classical iterative solvers~\cite{kopanivcakova2024deeponet,zhang2022hybrid}. 
This is intuitively effective because classical solvers are well known for handling high-frequency errors, but struggle with low-frequency modes.
In contrast, machine learning-based methods, such as NOs, exhibit the opposite behavior, due to a phenomenon known as the spectral bias of neural networks~\cite{rahaman2019spectral}). %By combining these two, we could obtain the best of both worlds. 

\subsection{Towards synthesizing meta-solvers}
Similar in spirit to the effective hybrid preconditioning strategy, the concept of \textit{meta-solvers} has been recently introduced in~\cite{lee2024automatic} by the authors. In~\cite{lee2024automatic}, two families of meta-solvers are parameterized and constructed by combining neural operators with classical relaxation iterative solvers and Krylov methods, respectively. The combination is then instantiated by varying the proportions between the two components. Additionally, advanced techniques, such as the multi-grid method, are also employed during the creation of meta-solvers. As a result, two families of meta-solvers are parametrized within two distinct multi-dimensional subspaces of different hyperparameters, corresponding to the relaxation-based and Krylov-based meta-solvers, respectively.  Given a specific element within these subspace, a meta-solver is then instantiated. 

Two natural questions arise regarding the meta-solver approach: Are the meta-solvers always better than neural operators and classical iterative solvers from all perspectives (e.g, computational time, error, number of iterations, etc.)? And if so, what is the maximum ``speed-up'' they can achieve in each of the aforementioned perspectives. 
In~\cite{lee2024automatic}, the authors propose a multi-objective optimization and Pareto-optimality methodology, based on evaluating the parametrized performance matrix, to identify  the optimal meta-solver. 
The studies and performance matrix evaluation are conducted in the context of solving Poisson equations, as an academic example. 
The illustrative numerical results demonstrate consistent improvements across all the aforementioned performance criteria, when compared to existing numerical solvers. Moreover, the results show convincing generalizability on different problems, as well as the extension of the parameterization space and performance matrix.

\subsection{Contributions and main results}

In this work, we develop a meta-solver approach to address one of the most challenging tasks: solving the scattering problem (see, for instance,~\cite{bayliss1980radiation, turkel1998absorbing, turkel2013compact, ogilvy1987wave, voronovich2013wave}) with various scatterer geometries in 3D settings. 
The governing equation is the Helmholtz equation, which is generally not positive-definite and involves complex-valued boundary conditions that destroy symmetry. 

\begin{figure}[H]
\includegraphics[width=0.99\textwidth]{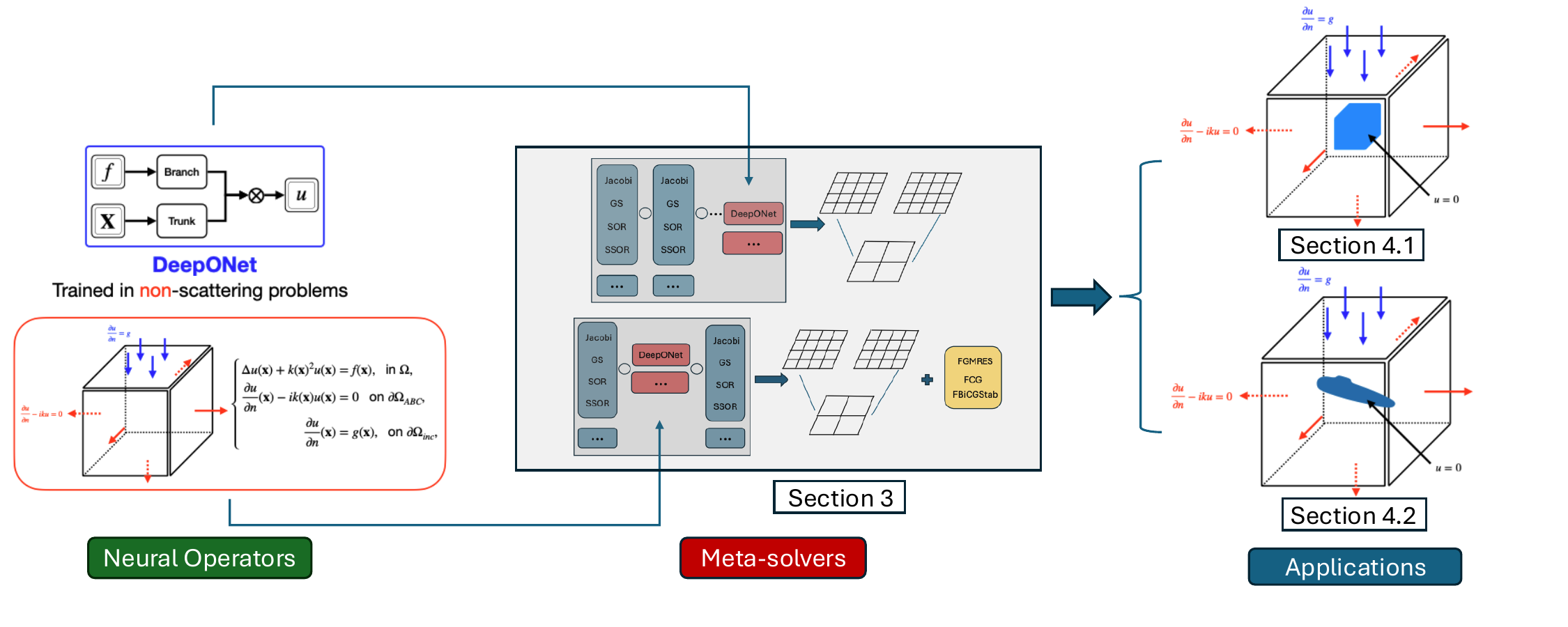}
\caption{A sketch of the meta-solver approach for 3D scattering problems. The DeepONet is first trained in a non-scattering context. Relaxation-based and Krylov-based meta-solvers are then developed by combining classical iterative solvers with DeepONet. These meta-solvers are tested on cube scattering and submarine-like scattering problems.}
\label{sketch_intro}
\end{figure}

Mathematically, we 
formulate the problem with an arbitrary scatterer inside the domain, an incoming wavefront from one of the boundary edges, and a Sommerfeld-type reflecting boundary condition. 
We propose leveraging the extrapolation capabilities of neural operators \cite{lu2022comprehensive, zhu2023reliable, kahana2023geometry}, particularly DeepONet, and using them to address scatterers with varying geometries. 
% We propose leveraging the extrapolation capabilities of neural operators \cite{lu2022comprehensive, zhu2023reliable, kahana2023geometry}, particularly DeepONet, and using them as \textit{foundation models} to address scatterers with varying geometries. 
The neural operators are trained on \textit{non-scattering} scenarios. 
Then, two families of meta-solvers are constructed, by combining DeepONet with either relaxation iterative solvers or Krylov solvers. 
Notice that the neural operators are applied independent of any specific scatterer information. 
We also explore the multi-grid techniques, in the creation of meta-solvers. 
A sketch of the methodology is shown in~\Cref{sketch_intro}. 

The meta-solvers are implemented and compared with standard relaxation and Krylov methods in scattering problems involving scatterers of various shapes, showing effectiveness, robustness (or shape-agnostic nature), and efficiency of this approach. 
In particular, we first consider a cube scatterer placed at the center of the domain. In that case, the relaxation methods diverge, while the DeepONet-relaxation methods converge, showing the neural operators' capability. Compare to classical Krylov methods, the DeepONet based meta-solvers are approximately 15 times faster in the sense of iterations, and 5 times faster in the sense of computational time. 
We then implemented a submarine-like scatter, and similar improvements are also observed. 
Finally, we demonstrate the generalizability of the proposed meta-solver approach, by considering various wave numbers, ranging from 6 to 36, for the submarine-like scatter.

The paper is organized as follows: in~\Cref{sec:2}, we present the problem formulation and in~\Cref{sec:3}, we present the methodology. In~\Cref{sec:4}, we present the computational results in 3D scattering problems, including cube and submarine shape scatterers, and in~\Cref{sec:5}, we conclude with a summary. In the Appendix, we include additional 2D results for different scatterers and details of the implementation.

\section{Problem Formulation}\label{sec:2}

\begin{figure}
     \centering
     \includegraphics[width=0.8\textwidth]{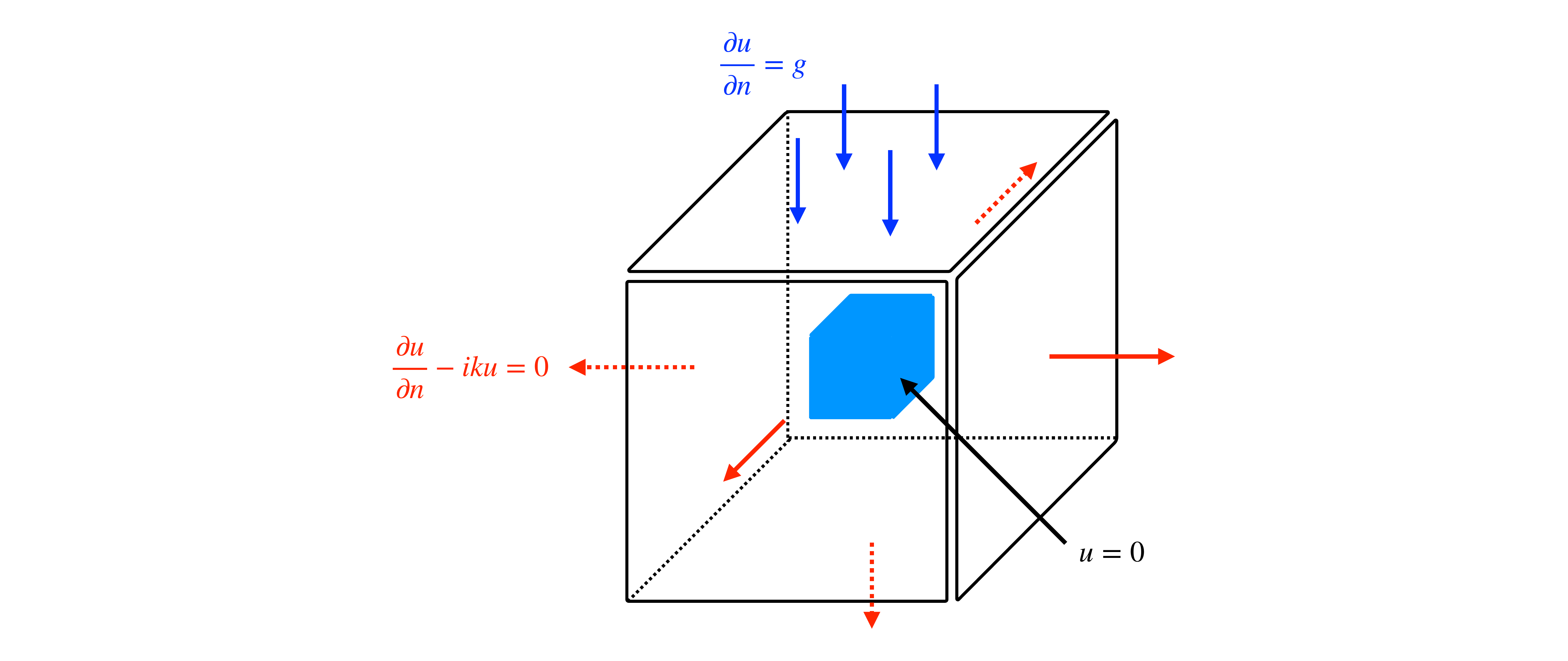}
     \caption{Setup of a standard scattering problem. The scatterer is illustrated by the shaded area in the middle of the domain. The upper boundary condition creates an incoming wavefront, while the other three boundary conditions are absorbing ones. Inside the domain, we solve the complex Helmholtz equation. The boundary of the scatterer is set as a reflecting boundary using a homogeneous Dirichlet condition.}
     \label{fig:1}
\end{figure}

The problem consists of a domain $\Omega$, with a scatterer inside the domain $\Omega_{scat}\subset \Omega$. The scatterer is assumed to be perfectly reflecting and have a homogeneous Dirichlet boundary condition. An incoming wave, denoted as $g$, is projected onto the domain from one of the sides, denoted as $\partial \Omega_{inc}$ (in this work, the top boundary), where a Neumann boundary condition is imposed.
The other boundaries, denoted as $\partial \Omega_{ABC}$, are open boundaries, using a Sommerfeld absorbing boundary condition~\cite{sommerfeld1949partial, zarmi2013general, BGT1982, gan1993finite}.
The general complex Helmholtz problem formulation for this setup is given by:
\begin{equation}
\label{eq:problem}
\left\{
    \begin{aligned}
        \Delta u(\mathbf{x}) + k(\mathbf{x})^2u(\mathbf{x}) &= f(\mathbf{x}), \quad \mathbf{x} \in \Omega\subset\mathbb{R}^{3}, \\
        \frac{\partial u}{\partial n}(\mathbf{x}) - ik(\mathbf{x})u(\mathbf{x}) &= 0, \quad\quad \ \ \mathbf{x} \in \partial \Omega_{ABC}, \\
        \frac{\partial u}{\partial n}(\mathbf{x}) &= g(\mathbf{x}), \quad\, \mathbf{x} \in \partial \Omega_{inc}, \\
        u(\mathbf{x}) &= 0, \quad\quad\ \ \mathbf{x} \in \Omega_{scat} \subset \Omega,
    \end{aligned}
\right.
\end{equation}
where $k(\mathbf{x})$ denotes the real space-dependent wave number and $f(\mathbf{x})$ denotes the complex space-dependent forcing term.
An illustration of the problem in a three-dimensional cube is given in~\Cref{fig:1}, where the domain $\Omega$ is a cube, and the forcing term is set to zero.
The dynamics of the system are introduced by the incoming wave $g(\mathbf{x})$~(assumed real).

In this paper, we focus on the scattering problem described by the complex Helmholtz equation~\eqref{eq:problem} in 3D, i.e., $\Omega = [0, 1]^{3}$. 
The domain $\Omega $ is discretized using a quasi-uniform tetrahedron mesh, as commonly employed in finite element method.
Without loss of generality, we set the forcing term $f$ to zero for the physical problem, in our experiments.
However, the forcing term is not zero when training the neural operators to stabilize and accelerate the iterative solver.
The implementation details will be presented in~\Cref{sec:3}. 
We also implemented our approach in various 2D scenarios, and the results are presented in~\ref{sec:app2d}.

%Recall that the GMRES~\cite{saad1986gmres} or BiCGStab~\cite{van1992bi} algorithms are typically used with a certain preconditioner since the problem~\eqref{eq:problem} is nonsymmetric.
%However, defining the preconditioner is not easy when we do not have any information about the scatterer.
%For instance, in real-world problems, the submarine causes scattering of the acoustic wave, where the source should be detected in real-time.
%Since the submarine's shape is usually nonconvex and multiply connected, designing the preconditioner is difficult.
%In addition, if the shape changes, the preconditioner might be redefined.

\section{Fast meta-solvers using pre-trained neural operators}
\label{sec:3}

In this section, we describe the fast meta-solvers approach for solving the scattering problem. 
In particular, we present the hybrid preconditioners that combine DeepONet with both relaxation-based methods and Krylov methods.
We then give the details of training the neural operators on a non-scattering problem. 

\subsection{Preconditioning the iterative solvers using neural operators}
\label{sec:hints}
\subsubsection{Relaxation-based  meta-solvers}

\begin{figure}
\centering
\includegraphics[width=1\textwidth]{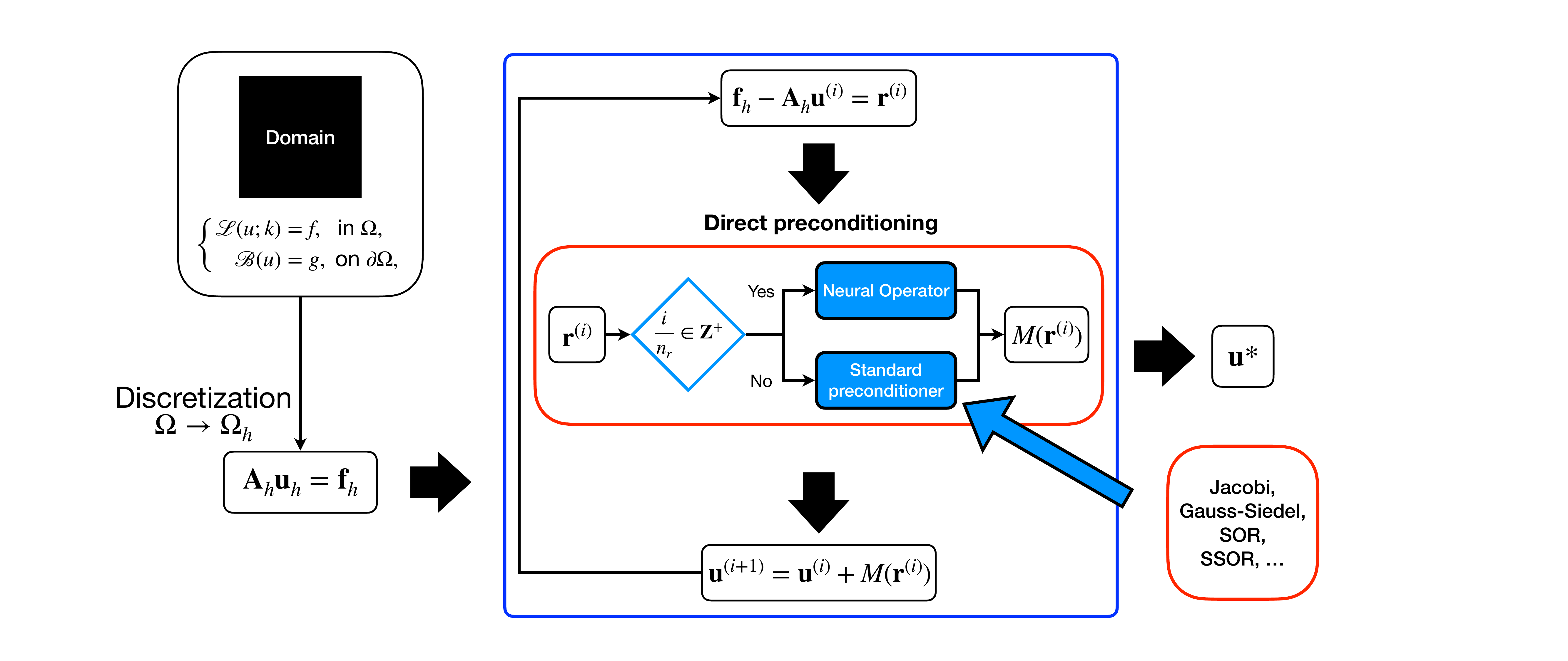}
\caption{Schematic view of the relaxation-based hybrid preconditioning strategy.
The DeepONet is chosen as the backbone neural operator for the hybrid preconditioning strategy.
Jacobi, Gauss-Seidel, SOR, and SSOR are commonly used as the standard preconditioner.}
\label{fig:relax}
\end{figure}
Let us assume that the given problem~\eqref{eq:problem} is discretized as a linear system $\mathbf{A}\mathbf{u}=\mathbf{f}$, using the finite element method.
The $i$-th iterate $\mathbf{u}^{(i)}$ of the standard relaxation method is regarded as the preconditioned Richardson iteration and is given by
\begin{equation}
\label{eq:relaxation}
\begin{split}
\mathbf{r}^{(i)} &= \mathbf{f} - \mathbf{A}\mathbf{u}^{(i)}, \\
    \mathbf{u}^{(i+1)} &= \mathbf{u}^{(i)} + \mathbf{M}(\mathbf{r}^{(i)}),
\end{split}
\end{equation}
where $\mathbf{M}$ denotes the preconditioner. 
We first consider the relaxation-based hybrid preconditioners. 
Indeed, for the relaxation methods, 
taking $\mathbf{M}$ as the diagonal part of $\mathbf{A}$ or the lower triangular part of $\mathbf{A}$ leads to the Jacobi and GS methods, respectively.
Under the multiplicative subspace correction framework~\cite{xu1992iterative}, the hybrid preconditioned Richardson iteration introduced in~\cite{zhang2022hybrid}, which further improves the iteration~\eqref{eq:relaxation}, can be written as
\begin{equation}
\label{eq:hybrid}
\begin{split}
\mathbf{r}^{(i)} &= \mathbf{f} - \mathbf{A}\mathbf{u}^{(i)}, \\
\mathbf{u}^{(i+1/2)} &= \mathbf{u}^{(i)} + \mathbf{M}_{1}(\mathbf{r}^{(i)}), \\
\mathbf{r}^{(i+1/2)} &= \mathbf{f} - \mathbf{A}\mathbf{u}^{(i+1/2)}, \\
\mathbf{u}^{(i+1)} &= \mathbf{u}^{(i+1/2)} + \mathbf{M}_{2}(\mathbf{r}^{(i+1/2)}), \\
\end{split}
\end{equation}
where $M_{1}$ 
%and $M_{2}$ 
represents $n_{r}$ steps of the relaxation method, 
$M_{2}$ represents the inference through neural operators, where $n_{r}$ is the ratio between the relaxation method and the NO, which is manually determined. 
A sketch of the construction of relaxation-based meta-solvers is presented in~\Cref{fig:relax}. 
%$n_{r}$ steps of the relaxation method and inference through the NO.
%$n_{r}$ represents the ratio between the relaxation method and the NO, which is manually determined.
%The practical details regarding the operator $\mathbf{M}_{2}$ can be found in~\cite{zhang2022hybrid}.

\subsubsection{Krylov-based meta-solvers}
\begin{figure}
\centering
\includegraphics[width=1\textwidth]{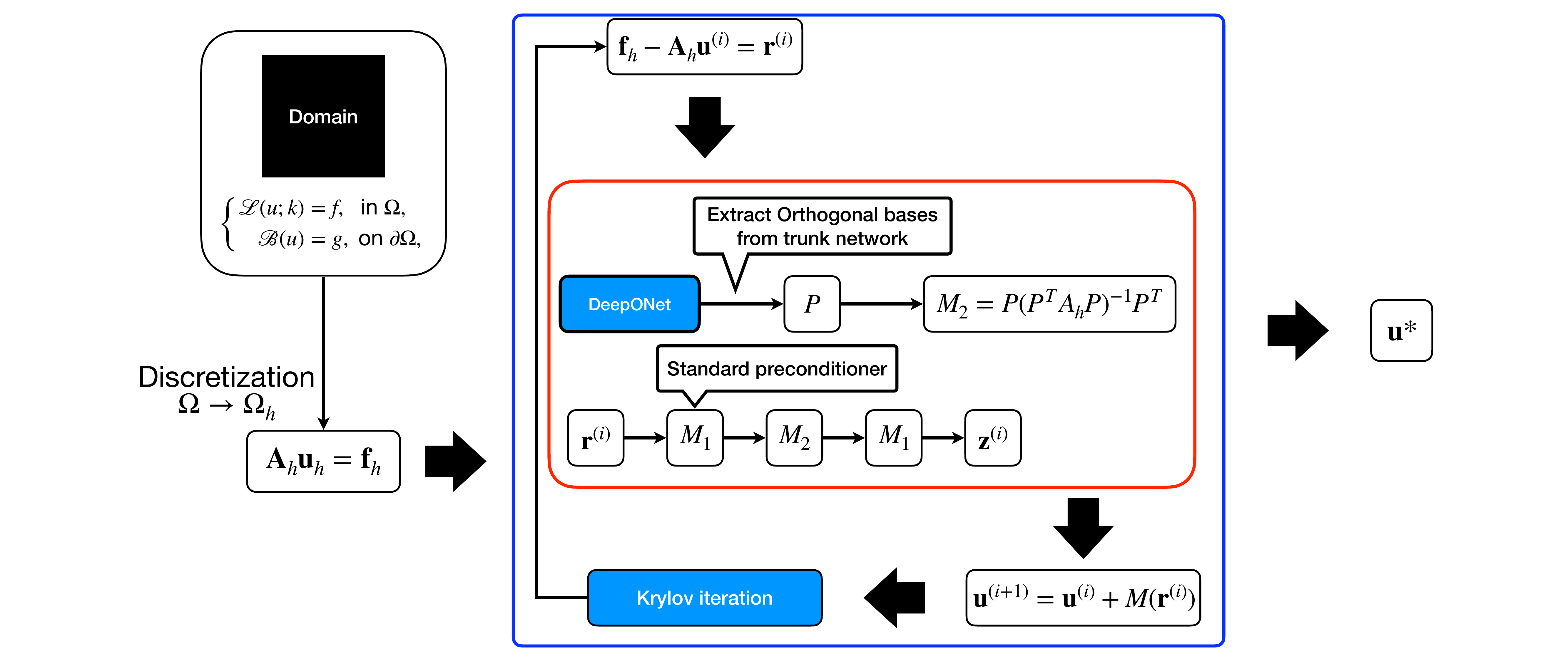}
\caption{Schematic view of the Krylov-based hybrid preconditioning strategy.
The DeepONet is chosen as the backbone neural operator for the hybrid preconditioning strategy.
% The DeepONet is chosen as the foundation model for the hybrid preconditioning strategy.
Since the inference of the neural operator is a nonlinear operation, direct preconditioning cannot be used as the preconditioner of the Krylov method. Instead, the trunk basis (TB) approach is employed.}
\label{fig:krylov}
\end{figure}
System~\eqref{eq:hybrid} is typically well defined for relaxation-based hybrid preconditioners. 
However, this approach cannot be directly used to precondition the Krylov method because the inference through the neural operator is a nonlinear operation. 
To overcome this challenge, the trunk basis~(TB) approach~\cite{kopanivcakova2024deeponet} was proposed. %, where the DeepONet is employed as the backbone of the NO.
In this approach, the prolongation and restriction operators $\mathbf{P}, \mathbf{R}$ are constructed by extracting the basis functions from the trunk network of the pre-trained DeepONet. The mathematical details regarding the prolongation operator $\mathbf{P}$ can be found in~\ref{sec:tb_approach}.
A sketch of the construction of Krylov-based meta-solvers is presented in~\Cref{fig:krylov}. 

\subsection{Training neural operators on a non-scattering problem}
Intuitively, for the scattering problem described in \eqref{eq:problem}, the neural operator used as a preconditioner for the iterative solvers should be trained to learn the solution operator $\mathcal{G}$ of the same scattering problem, i.e., \eqref{eq:problem} with the same scatterer $\Omega_{scat}$. This operator maps the wave number $k$, the forcing term $f$, and the incoming wave $g$ to the sought solution. Here, $k, f,$ and $g$ are space-dependent functions. While this approach can effectively solve scattering problems involving the same scatterer but varying $k, f$ and $g$, it requires exact prior knowledge of the scatterer before training the NO. Additionally, a new NO must be trained whenever a new scatterer, denoted by $\tilde{\Omega}_{scat}$, is introduced, ensuring the NO utilized to precondition the iterative solvers accurately approximates the solution operator of the updated scattering problem with $\tilde{\Omega}_{scat}$.
This constraint significantly limits the versatility and applicability of the approach, particularly in addressing diverse scatterer configurations, e.g., different geometries of the scatterer.

In this work, we leverage the extrapolation capability of NOs, specifically DeepONets,  %~\cite{lu2021learning,lu2022comprehensive,benitez2024out, zhang2022hybrid,kahana2023geometry,versanogeometry} 
to precondition iterative solvers for scattering problems by employing NOs trained on non-scattering problems. Specifically, the NO used in \eqref{eq:prolongation} approximates the solution operator of the Helmholtz equation defined in \eqref{eq:problem} with $\Omega_{scat}=\emptyset$ (we denote this operator as $\mathcal{G}_{non}$). This approach, which utilizes NOs trained to approximate $\mathcal{G}_{non}$ for solving scattering problems, offers two key advantages:
\begin{enumerate}
    \item \textbf{Simplicity of approximation}: $\mathcal{G}_{non}$ is generally easier for NOs to approximate, as non-scattering problems are often less complex than scattering problems.
    \item \textbf{Versatility across scatterer geometries}: Once trained, the NO approximating $\mathcal{G}_{non}$ can be applied to solve scattering problems with varying scatterer geometries without any retraining. This significantly enhances the generalizability and efficiency of the proposed approach in handling scattering problems.
\end{enumerate}
Since the hybridization strategy is based on preconditioned Richardson iteration, the performance of the solvers is determined by how accurately the preconditioner approximates the inverse of the stiffness matrix $\mathbf{A}^{-1}$.
For example, the SOR preconditioner approximates $\mathbf{A}^{-1}$ using the inverse of the lower triangular part of $\mathbf{A}$.
In this point of view, even if the NO is trained on a non-scattering problem, the pre-trained NO closely approximates $\mathbf{A}^{-1}$ except the obstacle region $\Omega_{scat}$.
That is, the pre-trained NO is independent of any specific scatterer information and can, in principle, be applied to problems involving arbitrary scatterer geometries.
Furthermore, it is enough for NO to be trained on homogeneous problems because NO acts on the residual of PDE.
The mathematical details of this property are described in~\ref{sec:homogenuous}.
% We refer to such NOs as \textit{foundation models}, as they are independent of any specific scatterer information and can, in principle, be applied to problems involving arbitrary scatterer geometries.
Numerical results presented in~\Cref{sec:4} empirically demonstrate the effectiveness of this scatterer-agnostic NO in solving 3D scattering problems.

\section{Results}
\label{sec:4}
We present the test results of using meta-solvers to solve the scattering problem described by the Helmholtz equation~\eqref{eq:problem} with a complex absorbing boundary condition.
The FEniCSx library~\cite{BarattaEtal2023} is used to discretize the scattering problem.
All meta-solvers were implemented by combining PETSc~\cite{osti_2205494} with PyTorch~\cite{paszke2019pytorch}.
To construct the meta-solvers using a hybrid preconditioning strategy, we use the DeepONet as the backbone neural operator and the Jacobi, Gauss-Seidel, SOR, SSOR, GMRES, and BiCGStab as the backbone iterative solver, respectively.
The conjugate gradient method cannot be employed since the problem~\eqref{eq:problem} is nonsymmetric and not positive definite.
All experiments were conducted using computational resources and services at the Center for Computation and Visualization, Brown University.
Each computing node is equipped with an AMD EPYC 9554 64-Core Processor~(256GB) and an NVIDIA L40S GPU~(48GB).

\begin{figure}
	\centering
     \begin{subfigure}[b]{0.3\textwidth}
         \centering
         \includegraphics[width=\textwidth]{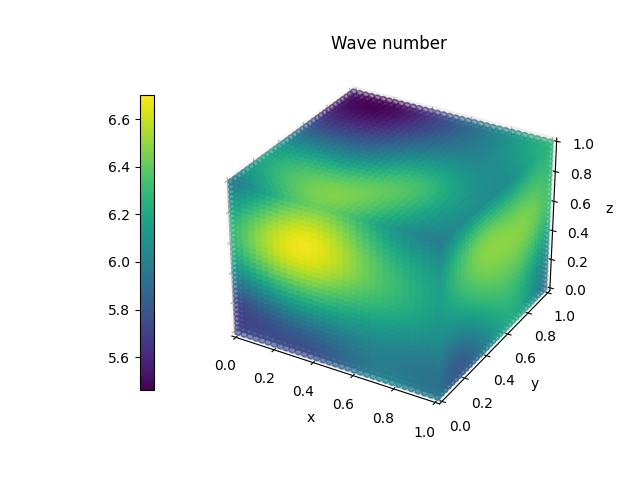}
         \caption{Wave number $k(\mathbf{x})$.}
     \end{subfigure}
     \begin{subfigure}[b]{0.6\textwidth}
         \centering
         \includegraphics[width=\textwidth]{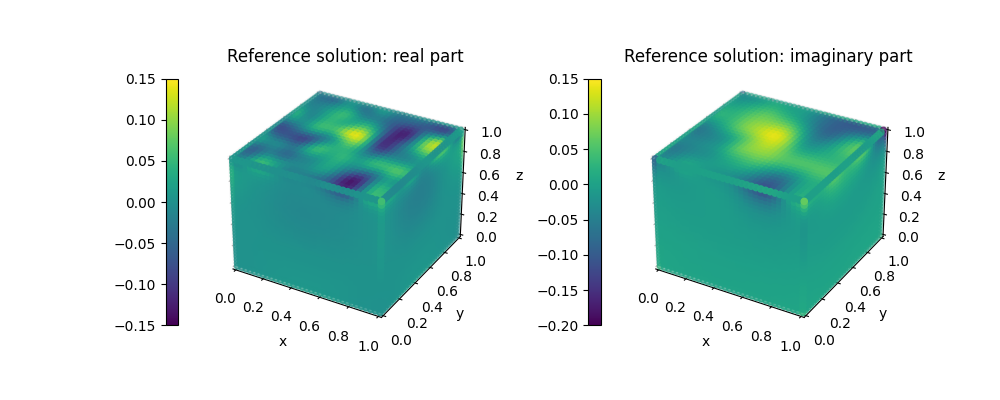}
         \caption{Reference solution.}
     \end{subfigure}
     \begin{subfigure}[b]{1\textwidth}
    \includegraphics[width=\textwidth]{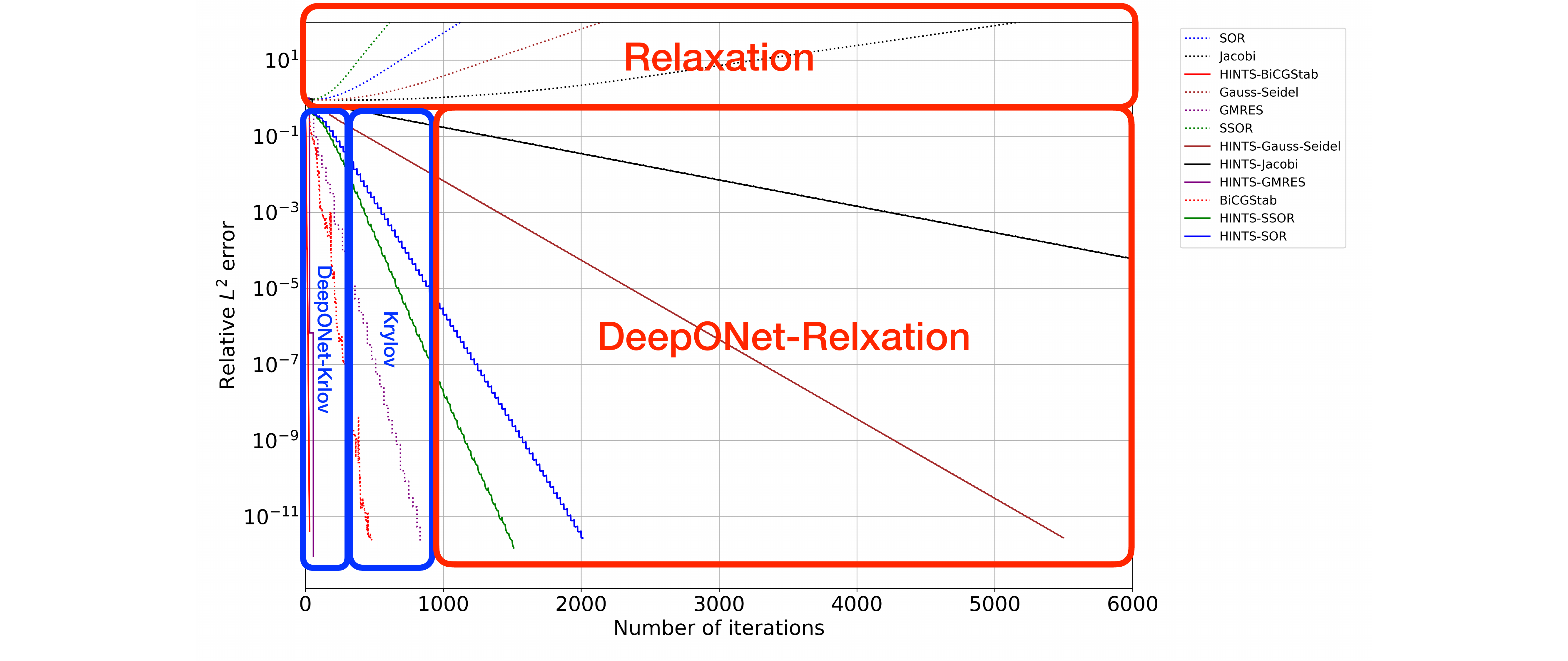}
    \caption{Relative $L^{2}$ error w.r.t. the number of iterations for various solvers.}
    \end{subfigure}
	\caption{(a) shows the 3D wave number $k(\mathbf{x})$ and (b) shows the reference solution. (c) shows the relative $L^{2}$ error of various classical solvers~(dotted line) and meta-solvers~(solid line) solving the 3D scattering problem on $[0,1]^{3}$ with a cubic scatterer (with size $0.125^{3}$) placed at the center. We discretize the computational domain with a $33 \times 33 \times 33$ quasi-uniform tetrahedron mesh, while the NO in HINTS is trained on 3D non-scattering problems with a $17 \times 17 \times 17$ quasi-uniform tetrahedron mesh. The number of steps of the relaxation is $n_{r}=3$ and the number of trunk-basis is $p=32$.}
	\label{fig:cube}
\end{figure}

\begin{figure}
	\centering
     \begin{subfigure}[b]{\textwidth}
         \centering
         \includegraphics[width=\textwidth]{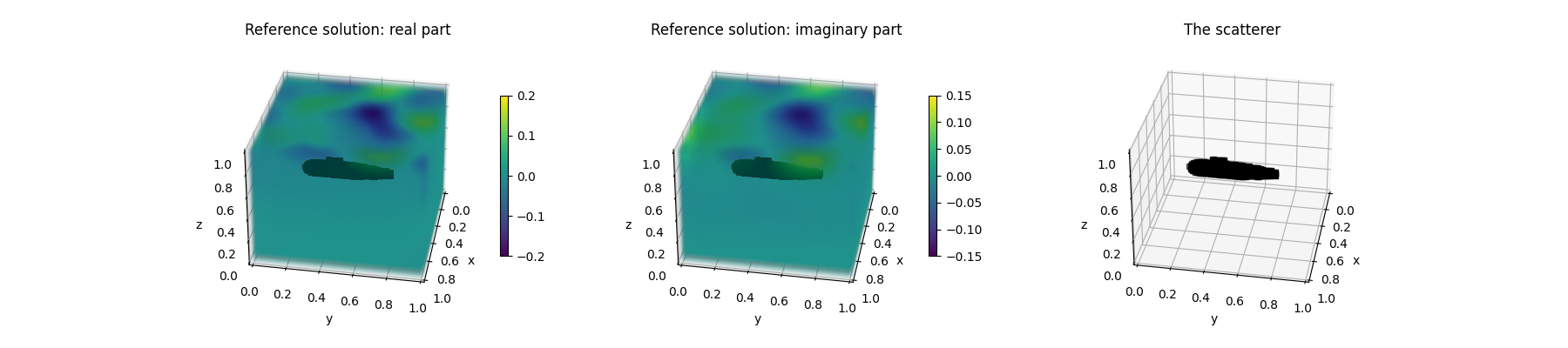}
         \caption{Reference solution and the submarine geometry~(rightmost)}
     \end{subfigure}
     \begin{subfigure}[b]{1\textwidth}
\includegraphics[width=\textwidth]{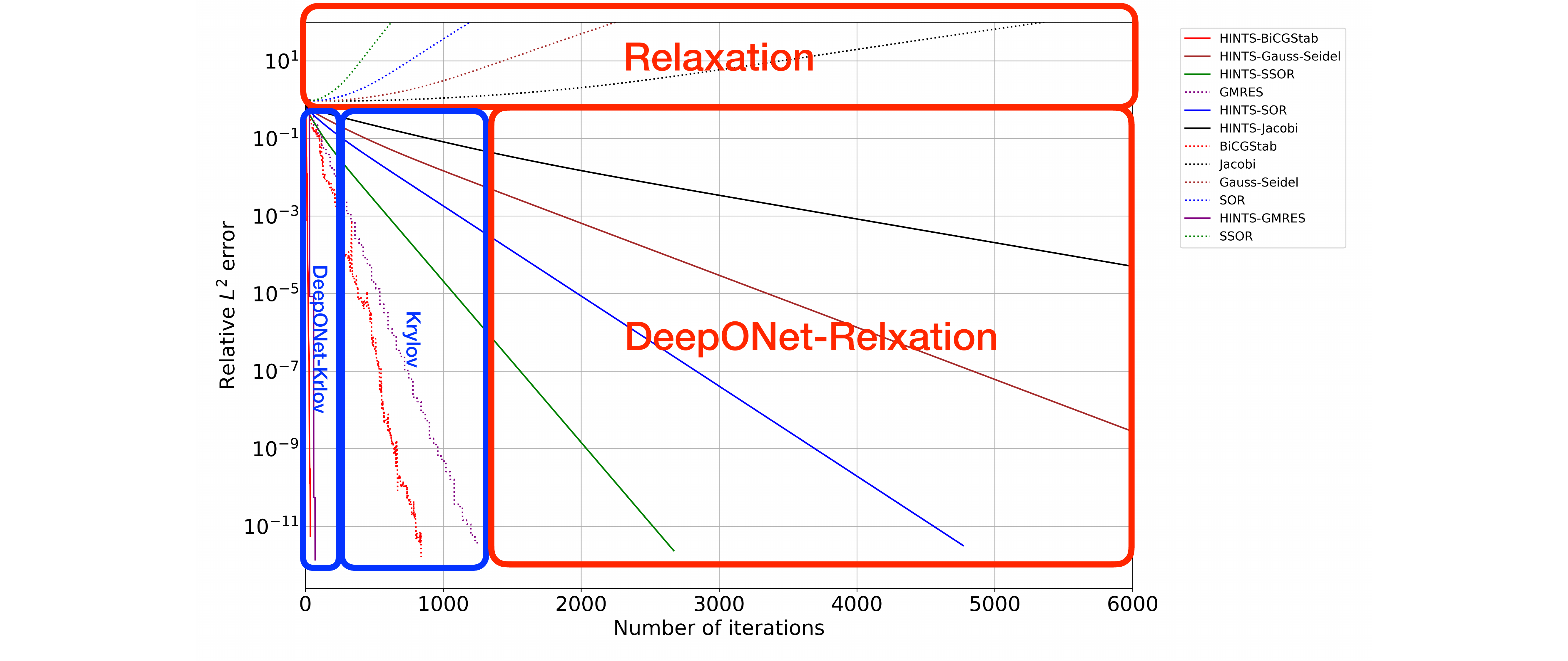}
\caption{Relative $L^{2}$ error w.r.t. the number of iterations for various solvers.}
\end{subfigure}
	\caption{(a) The submarine scatterer is rescaled to fit the domain. (b) shows the relative $L^{2}$ error of various classical solvers~(dotted line) and meta-solvers~(solid line) solving the 3D scattering problem on $[0,1]^{3}$ with a submarine scatterer. We discretize the computational domain with a $40 \times 40 \times 40$ quasi-uniform tetrahedron mesh, while the NO in HINTS is trained on 3D non-scattering problems with a $17 \times 17 \times 17$ quasi-uniform tetrahedron mesh.
    The number of steps of the relaxation is $n_{r}=3$ and the number of trunk-basis is $p=32$.}
	\label{fig:submarine}
\end{figure}

\subsection{A cubic-shaped scatterer}

To demonstrate the performance of proposed meta-solvers, we solve the 3D scattering problem on a cubic domain $\Omega:= [0, 1]^{3}$.
The Sommerfeld boundary condition is imposed on the front, back, left, right, and bottom boundaries of $\Omega$, a Neumann boundary condition with an incoming wave $g(\mathbf{x})$ is imposed on the top boundary, and a Dirichlet boundary condition is imposed on the scatterer.
The wave number $k(\mathbf{x})$ (displayed in~\Cref{fig:cube}(a)) and the incoming wave $g(\mathbf{x})$ are randomly sampled from a 3D Gaussian random field~(GRF) with mean $6.0$ and radial basis function~(RBF) kernel with $l=0.3$ and $s=0.2$ and the 2D GRF with mean zero and kernel function with $l=0.1$ and $s=1.0$, respectively.
The forcing term $f(\mathbf{x})$ is set to zero.
The relative $L^{2}$ error is the $L^{2}$ norm of the difference between the solution from direct solver and the solution from iterative solvers.

We first consider a cube scatterer of size $0.125^{3}$, placed at the center of the domain.
In this case, we discretize the domain with a $33 \times 33 \times 33$ quasi-uniform triangular mesh and solve the scattering problem using various classical solvers and the proposed meta-solvers.
The reference solution is computed by a direct solver using the LU decomposition and is displayed in~\Cref{fig:cube}(b).
In the first plot of~\Cref{fig:cube}(c), every relaxation method~(dotted line) diverges while the DeepONet-relaxation method~(solid line) converges.
This implies that the preconditioning using the neural operator makes the relaxation method solve the scattering problems formulated by the complex Helmholtz equation.
As is well known, the Krylov methods~(GMRES and BiCGStab) can solve the scattering problem, but they require many iterations.
However, using the DeepONet-Krylov method, we significantly reduce the number of iterations for GMRES from $832$ to $58$ and for BiCGStab from $483$ to $30$, respectively.
The DeepONet is trained on 3D non-scattering problems and a $17 \times 17 \times 17$ quasi-uniform tetrahedron mesh.
% The DeepONet is trained on 3D non-scattering problems and a $17 \times 17 \times 17$ quasi-uniform tetrahedron mesh, which can be regarded as the foundation model.

\subsection{A submarine-shaped scatterer}

We next solve the 3D scattering problem with a submarine-like scatterer. 
The submarine has been modeled with Computer Aided Design~(CAD) software.
The model is not of a classified operational submarine but captures the general shape.
%The code used to parse the model from the CAD software can be used for the classified models.
The model features the body of the submarine, as well as a tower on top of it.
Focusing only on the NO part of the meta-solvers, we do not use a mesh to represent the submarine. Instead, we use a stair-casing technique with a uniform grid, assigning the value $1$ to nodes within the submarine geometry and $0$ otherwise.
The NO interpolates the space and reduces the low-frequency error modes. This is a sufficient assumption to achieve satisfactory results in this case.
Thus, the NO scales with the size of the problem of interest. We scale the submarine to fit the domain $[0, 1]^3$; see~\Cref{fig:submarine}(a).
We discretize the domain with a $40 \times 40 \times 40$ quasi-uniform tetrahedron mesh.
The wave number and the incoming wave are the same as in the previous case but interpolated linearly for the new mesh.
We employ the same NO as the backbone neural operator for meta-solvers, trained on 3D non-scattering problems with the $17 \times 17 \times 17$ quasi-uniform tetrahedron mesh.
% We employ the same NO as the foundation model for meta-solvers, trained on 3D non-scattering problems with the $17 \times 17 \times 17$ quasi-uniform tetrahedron mesh.
The results for various meta-solvers to solve the scattering problem with the submarine geometry are shown in~\Cref{fig:submarine}(b).
Compared to the cube scatterer case described in~\Cref{fig:cube}(b), the number of iterations required for convergence increases in all cases.
However, it is consistent that every DeepONet-Krylov method shows the best results. 

\begin{table}
\centering
\tiny
\begin{tabular}{@{}c|c|c|c|c|c|c|c|c|c|c|c|c@{}}
\toprule
\multirow{2}{*}{$k$} & \multicolumn{3}{c|}{BiCGStab} & \multicolumn{3}{c|}{ILU-BiCGStab} & \multicolumn{3}{c|}{SOR-BiCGStab} & \multicolumn{3}{c}{DeepONet-BiCGStab} \\ \cmidrule(l){2-13} 
 & Error & iters. & Time~(s) & Error & iters. & Time~(s) & Error & iters. & Time~(s) & Error & iters. & Time~(s) \\ \midrule
6 & 1.78e-12 & 758 & 2.62 & 1.79e-12 & 99 & 0.65 & 1.12e-12 & 153 & 1.07 & 4.75e-12 & 36 & 0.43 \\
12 & 1.46e-12 & 2178 & 7.48 & 1.96e-12 & 219 & 1.50 & 1.74e-12 & 355 & 2.59 & 3.40e-12 & 84 & 0.91 \\
18 & 1.14e-12 & 1922 & 6.51 & 1.20e-12 & 257 & 1.60& 1.38e-12 & 420 & 3.03 & 7.61e-13 & 171 & 1.69 \\
24 & 1.14e-12 & 1716 & 5.53 & 1.50e-12 & 307 & 1.89 & 1.61e-12 & 426 & 2.90 & 9.22e-13 & 219 & 2.23 \\
30 & 6.79e-13 & 1635 & 4.87 & 2.31e-12 & 383 & 2.18 & 5.80e-13 & 467 & 3.05 & 1.60e-12 & 282 & 2.91 \\
36 & 1.26e-12 & 1308 & 4.18 & 2.82e-12 & 466 & 3.03 & 2.013e-12 & 430 & 2.93 & 1.94e-12 & 397 & 4.25 \\ \bottomrule
\end{tabular}
    \caption{The relative $L^{2}$ error and the number of iterations with respect to the wave number $k$. Note that $M$-BiCGStab denotes the BiCGStab method with the preconditioner $M$. The relative $L^{2}$ error is the $L^{2}$ norm of the difference between the solution from direct solver and the solution from iterative solvers. The DeepONet for the hybrid preconditioner is trained at the wave number $k=6$. 
    % Note that the number of relaxation steps for the neural operator and the number of trunk-basis are $1$ and $32$, respectively.
    The ratio between the relaxation and neural operator is $1$ and the number of trunk-basis is $p=32$.
    }
    \label{tab:wave}
\end{table}

\subsection{Generalization to different wave numbers}

To demonstrate the generalizability of the proposed hybrid preconditioned solvers, we consider various wave numbers $k=6, 12, 18, 24, 30$, and $36$ for the submarine scatterer experiment.
When the wave number is large, the induced linear system loses its ellipticity, which implies that the iterative solver requires many iterations.
For this experiment, we chose the BiCGStab as the baseline, which showed the best performance on the scattering problem with the submarine scatterer.
In addition, we compare the performance with commonly used traditional preconditioners such as Incomplete LU decomposition~(ILU) and Successive Over-Relaxation~(SOR).
To compare the performance of each solver, we measure the number of iterations and computation time and compute the relative $L^{2}$ error between the solution from the direct solver and the solution from each iterative solver.
The performance comparison of the various preconditioned BiCGStab is reported in~\Cref{tab:wave}.
The DeepONet-BiCGStab shows the best number of iterations compared to others.
The DeepONet is implemented by Python, but the other preconditioners are implemented by C/C++.
Implementing DeepONet in C/C++ may lead to further improvements in computation time.
Even though the DeepONet used for hybrid preconditioned BiCGStab was trained at sampled small wave numbers $\mathbb{E}_{\mathbf{x}} k(\mathbf{x})=6$, the hybrid preconditioner can be employed for the problem with high wave numbers.

\begin{figure}
	\centering
    \includegraphics[width=\textwidth]{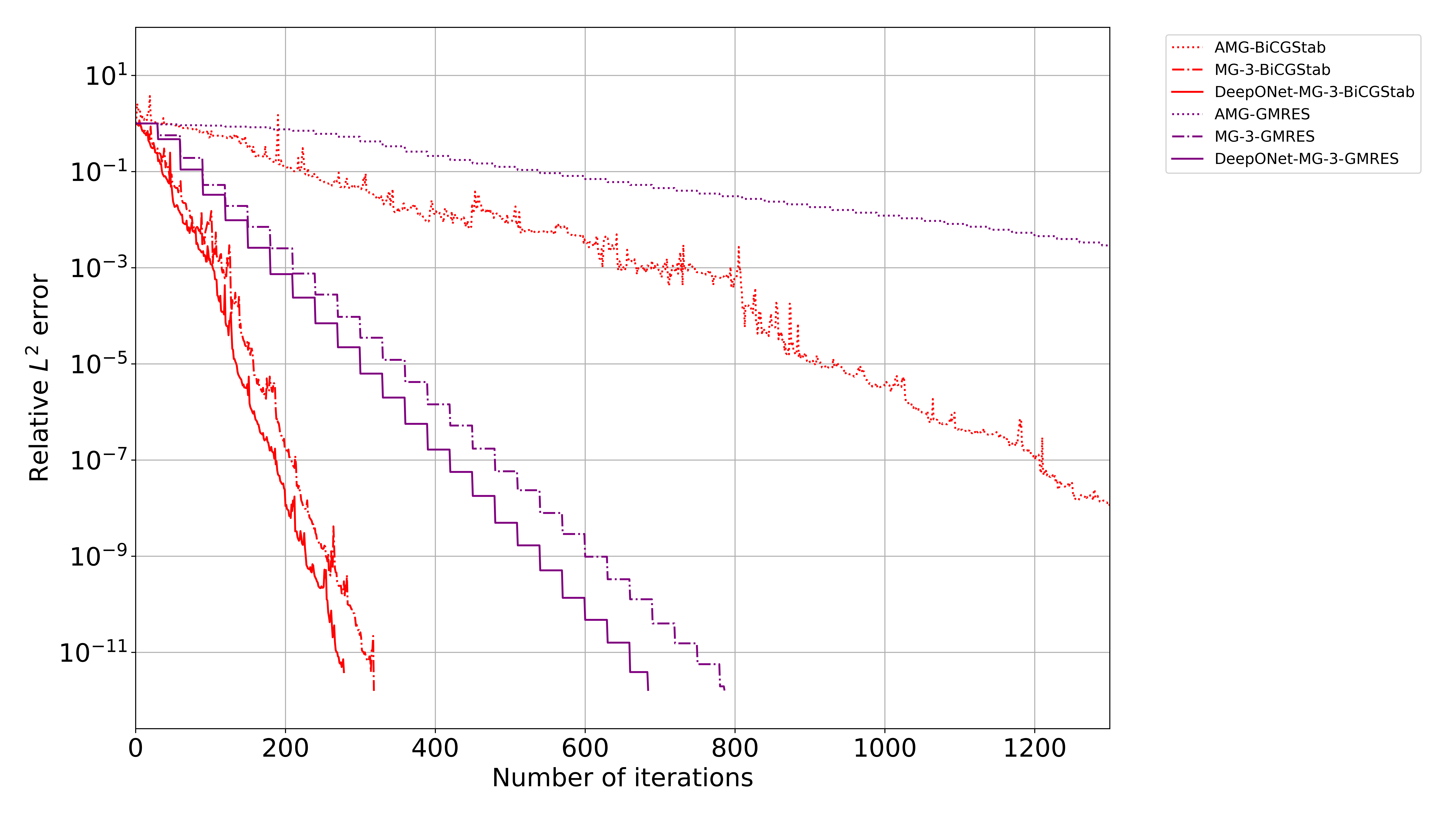}
    \caption{The decay of the relative $L^{2}$ error with respect to the number of iterations for the high wave number $k=36$. 
    For AMG-Krylov, the PETSc GAMG preconditioner is used.
    DeepONet-MG-Krylov denotes the Krylov method accelerated by multigrid with the DeepONet-Jacobi smoother. For MG-Krylov, the Jacobi smoother is used. The ratio between the relaxation and neural operator is $2$ and the number of trunk-basis is $p=32$.}
	\label{fig:mg}
\end{figure}

Even though the DeepONet-Relaxation method is slower than the standard Krylov or DeepONet-Krylov, we can utilize it as a smoother for a multigrid framework to accelerate the convergence of the Krylov methods.
\Cref{fig:mg} compares the standard multigrid~(MG) preconditioner with a Jacobi smoother, the algebraic multigrid~(AMG) preconditioner in PETSc, and the DeepONet-MG preconditioner with a Jacobi smoother.
The DeepONet enhanced MG preconditioner required fewer iterations than the standard MG preconditioner.
The wave number $k$ is $36$, and the number of smoothing steps is $5$ for MG-Krylov.
For DeepONet-MG-Krylov, we use the $2-1-2$ smoothing strategy, which eventually has the same number of smoothing steps as MG-Krylov.

\section{Summary}
\label{sec:5}
% We proposed fast meta-solvers for the 3D scattering problem described by the Helmholtz equation in an exterior domain with a complex absorbing boundary condition generated by utilizing the hybrid preconditioner with a foundation model.
We proposed fast meta-solvers for the 3D scattering problem described by the Helmholtz equation in an exterior domain with a complex absorbing boundary condition generated by utilizing the hybrid preconditioner with DeepONet trained on non-scattering problem.
We considered cube-shaped and submarine-shaped scatterers, and observed that a hybrid preconditioning approach blending neural operators~(NOs), which leverages their spectral bias, can significantly accelerate the convergence of iterative solvers.
Specifically, the DeepONet~\cite{lu2021learning} is pre-trained as foundation NO, which implies that the DeepONet is trained in non-scattering problems.
We then solve scattering problems across diverse scatterer geometries and different wave numbers quickly and accurately.

% One possible limitation of using pre-trained NOs to solve partial differential equations~(PDEs) is that a new NO must be trained each time a new family of PDEs is solved.
% However, this can be also overcome by pre-training the NO for different PDEs and then using light transfer learning.
%For instance, when the size or shape of the scatterer changes in the scattering problem, the NO needs to be retrained from scratch or at least fine-tuned to provide accurate inferences for the new problem.
Since the contribution of NOs in this hybrid approach is primarily to decrease the errors of the low-frequency modes~\cite{zhang2022hybrid,kahana2023geometry}, the proposed solvers demonstrate resilience with respect to shape up to some limit in size.
In this work, we mainly focused on investigating the NO's extrapolation capability in tackling different scatterers.
% In this work, we mainly focused on investigating the NO's extrapolation capability in tackling different scatterers, which is the key feature of the foundation model.
Specifically, we first trained the NO on the non-scattering problem, described by the complex Helmholtz equation without the scatterer.
Then, we employed the pre-trained NO of the hybrid preconditioning framework to solve the scattering problem with the scatterer in different shapes, including a cube and a scaled submarine in 3D cases.
% Then, we employed the pre-trained NO as the foundation model of the hybrid preconditioning framework to solve the scattering problem with the scatterer in different shapes, including a cube and a scaled submarine in 3D cases.
We presented significant evidence in the main paper and in the Appendix showcasing the effectiveness of hybrid preconditioning in dealing with such challenges, demonstrating its extrapolation capability and versatility in addressing different scatterers.
%However, performance is degraded when a hybrid preconditioner is applied to solve the scattering problem when the NO is trained on non-scattering ones.
%As the problem with a scatterer becomes further removed from the case of no scattering, the convergence rate decreases and may even stop converging.
%This issue can be mitigated by training a NO on more similar scattering problems.

Another potential advancement of the present approach is to identify (one of) the optimal solvers for different scattering problems. In this context, one could explore tools from the field of optimization, that formulate the search for optimal solvers as a multi-objective optimization problem, addressing the various dimensional demands in the evaluation. One possible way is to implement the methodology in~\cite{lee2024automatic} recently introduced by the authors. This is of independent interest, and we consider it as a future work. 

\section*{Acknowledgements}

We acknowledge the support of the  DARPA project DIAL (HR00112490484) and the DOE-MMICS
SEA-CROGS DE-SC0023191 award.
Y.L. is supported in part by Basic Science Research Program through the National Research Foundation of Korea~(NRF) funded by the Ministry of Education~(No. RS2023-00247199).
G.E.K. is supported by the ONR Vannevar Bush Faculty Fellowship (N00014-22-1-2795). 
We also acknowledge support from Ansys Inc.
We thank Professor Sergey Petropavlovsky from HSE University for supplying the CAD model of the submarine.
Z.Z. thanks Dr. Enrui Zhang for helpful discussions.

%Bibliography
\newpage
\bibliographystyle{unsrt}  
\bibliography{references}  

\begin{thebibliography}{10}

\bibitem{saad2003iterative}
Yousef Saad.
\newblock {\em Iterative methods for sparse linear systems}.
\newblock SIAM, Philadelphia, 2003.

\bibitem{hadjidimos2000successive}
A~Hadjidimos.
\newblock Successive overrelaxation (sor) and related methods.
\newblock {\em Journal of Computational and Applied Mathematics}, 123(1-2):177--199, 2000.

\bibitem{saad1986gmres}
Youcef Saad and Martin~H Schultz.
\newblock {GMRES}: A generalized minimal residual algorithm for solving nonsymmetric linear systems.
\newblock {\em SIAM Journal on Scientific and Statistical Computing}, 7(3):856--869, 1986.

\bibitem{van1992bi}
Henk~A Van~der Vorst.
\newblock {Bi-CGSTAB}: A fast and smoothly converging variant of {Bi-CG} for the solution of nonsymmetric linear systems.
\newblock {\em SIAM Journal on Scientific and Statistical Computing}, 13(2):631--644, 1992.

\bibitem{karniadakis2021physics}
George~Em Karniadakis, Ioannis~G Kevrekidis, Lu~Lu, Paris Perdikaris, Sifan Wang, and Liu Yang.
\newblock Physics-informed machine learning.
\newblock {\em Nature Reviews Physics}, 3(6):422--440, 2021.

\bibitem{cuomo2022scientific}
Salvatore Cuomo, Vincenzo~Schiano Di~Cola, Fabio Giampaolo, Gianluigi Rozza, Maziar Raissi, and Francesco Piccialli.
\newblock Scientific machine learning through physics--informed neural networks: Where we are and what’s next.
\newblock {\em Journal of Scientific Computing}, 92(3):88, 2022.

\bibitem{toscano2024pinns}
Juan~Diego Toscano, Vivek Oommen, Alan~John Varghese, Zongren Zou, Nazanin Ahmadi~Daryakenari, Chenxi Wu, and George~Em Karniadakis.
\newblock {From PINNs to PIKANs: Recent advances in physics-informed machine learning}.
\newblock {\em Machine Learning for Computational Science and Engineering}, 1(1):1--43, 2025.

\bibitem{raissi2019physics}
Maziar Raissi, Paris Perdikaris, and George~E Karniadakis.
\newblock Physics-informed neural networks: A deep learning framework for solving forward and inverse problems involving nonlinear partial differential equations.
\newblock {\em Journal of Computational Physics}, 378:686--707, 2019.

\bibitem{cai2021fluid}
Shengze Cai, Zhiping Mao, Zhicheng Wang, Minglang Yin, and George~Em Karniadakis.
\newblock Physics-informed neural networks ({PINNs}) for fluid mechanics: A review.
\newblock {\em Acta Mechanica Sinica}, 37(12):1727--1738, 2021.

\bibitem{cai2021heat}
Shengze Cai, Zhicheng Wang, Sifan Wang, Paris Perdikaris, and George~Em Karniadakis.
\newblock Physics-informed neural networks for heat transfer problems.
\newblock {\em Journal of Heat Transfer}, 143(6):060801, 2021.

\bibitem{mao2020physics}
Zhiping Mao, Ameya~D Jagtap, and George~Em Karniadakis.
\newblock Physics-informed neural networks for high-speed flows.
\newblock {\em Computer Methods in Applied Mechanics and Engineering}, 360:112789, 2020.

\bibitem{shukla2025neurosem}
Khemraj Shukla, Zongren Zou, Chi~Hin Chan, Additi Pandey, Zhicheng Wang, and George~Em Karniadakis.
\newblock {NeuroSEM: A hybrid framework for simulating multiphysics problems by coupling PINNs and spectral elements}.
\newblock {\em Computer Methods in Applied Mechanics and Engineering}, 433:117498, 2025.

\bibitem{pang2019fpinns}
Guofei Pang, Lu~Lu, and George~Em Karniadakis.
\newblock {fPINNs}: Fractional physics-informed neural networks.
\newblock {\em SIAM Journal on Scientific Computing}, 41(4):A2603--A2626, 2019.

\bibitem{lee2024two}
Youngkyu Lee, Alena Kopani{\v{c}}{\'a}kov{\'a}, and George~Em Karniadakis.
\newblock Two-level overlapping additive {S}chwarz preconditioner for training scientific machine learning applications.
\newblock {\em arXiv preprint arXiv:2406.10997}, 2024.

\bibitem{linka2022bayesian}
Kevin Linka, Amelie Sch{\"a}fer, Xuhui Meng, Zongren Zou, George~Em Karniadakis, and Ellen Kuhl.
\newblock Bayesian physics informed neural networks for real-world nonlinear dynamical systems.
\newblock {\em Computer Methods in Applied Mechanics and Engineering}, 402:115346, 2022.

\bibitem{zhang2024discovering}
Zhen Zhang, Zongren Zou, Ellen Kuhl, and George~Em Karniadakis.
\newblock Discovering a reaction--diffusion model for {Alzheimer’s disease by combining PINNs} with symbolic regression.
\newblock {\em Computer Methods in Applied Mechanics and Engineering}, 419:116647, 2024.

\bibitem{mishra2021physics}
Siddhartha Mishra and Roberto Molinaro.
\newblock Physics informed neural networks for simulating radiative transfer.
\newblock {\em Journal of Quantitative Spectroscopy and Radiative Transfer}, 270:107705, 2021.

\bibitem{zou2023hydra}
Zongren Zou and George~Em Karniadakis.
\newblock Multi-head physics-informed neural networks for learning functional priors and uncertainty quantification.
\newblock {\em Journal of Computational Physics}, 531:113947, 2025.

\bibitem{misyris2020physics}
George~S Misyris, Andreas Venzke, and Spyros Chatzivasileiadis.
\newblock Physics-informed neural networks for power systems.
\newblock In {\em 2020 IEEE Power \& Energy Society General Meeting (PESGM)}, pages 1--5. IEEE, 2020.

\bibitem{zou2024correcting}
Zongren Zou, Xuhui Meng, and George~Em Karniadakis.
\newblock Correcting model misspecification in physics-informed neural networks {(PINNs)}.
\newblock {\em Journal of Computational Physics}, page 112918, 2024.

\bibitem{yin2023generative}
Minglang Yin, Zongren Zou, Enrui Zhang, Cristina Cavinato, Jay~D Humphrey, and George~Em Karniadakis.
\newblock A generative modeling framework for inferring families of biomechanical constitutive laws in data-sparse regimes.
\newblock {\em Journal of the Mechanics and Physics of Solids}, 181:105424, 2023.

\bibitem{sahli2020physics}
Francisco Sahli~Costabal, Yibo Yang, Paris Perdikaris, Daniel~E Hurtado, and Ellen Kuhl.
\newblock Physics-informed neural networks for cardiac activation mapping.
\newblock {\em Frontiers in Physics}, 8:42, 2020.

\bibitem{chen2024leveraging}
Paula Chen, Tingwei Meng, Zongren Zou, J{\'e}r{\^o}me Darbon, and George~Em Karniadakis.
\newblock Leveraging multitime {Hamilton--Jacobi PDEs} for certain scientific machine learning problems.
\newblock {\em SIAM Journal on Scientific Computing}, 46(2):C216--C248, 2024.

\bibitem{chen2023leveraging}
Paula Chen, Tingwei Meng, Zongren Zou, J{\'e}r{\^o}me Darbon, and George~Em Karniadakis.
\newblock Leveraging {Hamilton-Jacobi PDEs with time-dependent Hamiltonians} for continual scientific machine learning.
\newblock In {\em 6th Annual Learning for Dynamics \& Control Conference}, pages 1--12. PMLR, 2024.

\bibitem{henkes2022physics}
Alexander Henkes, Henning Wessels, and Rolf Mahnken.
\newblock Physics informed neural networks for continuum micromechanics.
\newblock {\em Computer Methods in Applied Mechanics and Engineering}, 393:114790, 2022.

\bibitem{zou2025learning}
Zongren Zou, Zhicheng Wang, and George~Em Karniadakis.
\newblock Learning and discovering multiple solutions using physics-informed neural networks with random initialization and deep ensemble.
\newblock {\em arXiv preprint arXiv:2503.06320}, 2025.

\bibitem{chiu2022can}
Pao-Hsiung Chiu, Jian~Cheng Wong, Chinchun Ooi, My~Ha Dao, and Yew-Soon Ong.
\newblock {CAN-PINN}: A fast physics-informed neural network based on coupled-automatic--numerical differentiation method.
\newblock {\em Computer Methods in Applied Mechanics and Engineering}, 395:114909, 2022.

\bibitem{zou2024leveraging}
Zongren Zou, Tingwei Meng, Paula Chen, J{\'e}r{\^o}me Darbon, and George~Em Karniadakis.
\newblock Leveraging viscous {Hamilton--Jacobi PDEs} for uncertainty quantification in scientific machine learning.
\newblock {\em SIAM/ASA Journal on Uncertainty Quantification}, 12(4):1165--1191, 2024.

\bibitem{kiyani2025optimizer}
Elham Kiyani, Khemraj Shukla, Jorge~F Urb{\'a}n, J{\'e}r{\^o}me Darbon, and George~Em Karniadakis.
\newblock Which optimizer works best for physics-informed neural networks and kolmogorov-arnold networks?
\newblock {\em arXiv preprint arXiv:2501.16371}, 2025.

\bibitem{lu2021learning}
Lu~Lu, Pengzhan Jin, Guofei Pang, Zhongqiang Zhang, and George~Em Karniadakis.
\newblock Learning nonlinear operators via {DeepONet} based on the universal approximation theorem of operators.
\newblock {\em Nature Machine Intelligence}, 3(3):218--229, 2021.

\bibitem{li2021fourier}
Zongyi Li, Nikola~Borislavov Kovachki, Kamyar Azizzadenesheli, Burigede liu, Kaushik Bhattacharya, Andrew Stuart, and Anima Anandkumar.
\newblock Fourier neural operator for parametric partial differential equations.
\newblock In {\em International Conference on Learning Representations}, 2021.

\bibitem{wang2021learning}
Sifan Wang, Hanwen Wang, and Paris Perdikaris.
\newblock Learning the solution operator of parametric partial differential equations with physics-informed {DeepONets}.
\newblock {\em Science Advances}, 7(40):eabi8605, 2021.

\bibitem{kovachki2023neural}
Nikola Kovachki, Zongyi Li, Burigede Liu, Kamyar Azizzadenesheli, Kaushik Bhattacharya, Andrew Stuart, and Anima Anandkumar.
\newblock Neural operator: Learning maps between function spaces with applications to {PDEs}.
\newblock {\em Journal of Machine Learning Research}, 24(89):1--97, 2023.

\bibitem{lu2022comprehensive}
Lu~Lu, Xuhui Meng, Shengze Cai, Zhiping Mao, Somdatta Goswami, Zhongqiang Zhang, and George~Em Karniadakis.
\newblock A comprehensive and fair comparison of two neural operators (with practical extensions) based on fair data.
\newblock {\em Computer Methods in Applied Mechanics and Engineering}, 393:114778, 2022.

\bibitem{ovadia2023vito}
Oded Ovadia, Adar Kahana, Panos Stinis, Eli Turkel, Dan Givoli, and George~Em Karniadakis.
\newblock Vito: Vision transformer-operator.
\newblock {\em Computer Methods in Applied Mechanics and Engineering}, 428:117109, 2024.

\bibitem{he2023novel}
Junyan He, Seid Koric, Shashank Kushwaha, Jaewan Park, Diab Abueidda, and Iwona Jasiuk.
\newblock Novel {DeepONet} architecture to predict stresses in elastoplastic structures with variable complex geometries and loads.
\newblock {\em Computer Methods in Applied Mechanics and Engineering}, 415:116277, 2023.

\bibitem{zou2023uncertainty}
Zongren Zou, Xuhui Meng, and George~Em Karniadakis.
\newblock Uncertainty quantification for noisy inputs--outputs in physics-informed neural networks and neural operators.
\newblock {\em Computer Methods in Applied Mechanics and Engineering}, 433:117479, 2025.

\bibitem{li2023phase}
Wei Li, Martin~Z Bazant, and Juner Zhu.
\newblock Phase-field {DeepONet}: Physics-informed deep operator neural network for fast simulations of pattern formation governed by gradient flows of free-energy functionals.
\newblock {\em Computer Methods in Applied Mechanics and Engineering}, 416:116299, 2023.

\bibitem{psaros2023uncertainty}
Apostolos~F Psaros, Xuhui Meng, Zongren Zou, Ling Guo, and George~Em Karniadakis.
\newblock Uncertainty quantification in scientific machine learning: Methods, metrics, and comparisons.
\newblock {\em Journal of Computational Physics}, 477:111902, 2023.

\bibitem{benitez2024out}
Jose Antonio~Lara Benitez, Takashi Furuya, Florian Faucher, Anastasis Kratsios, Xavier Tricoche, and Maarten~V de~Hoop.
\newblock Out-of-distributional risk bounds for neural operators with applications to the {Helmholtz} equation.
\newblock {\em Journal of Computational Physics}, page 113168, 2024.

\bibitem{li2023fourier}
Zongyi Li, Daniel~Zhengyu Huang, Burigede Liu, and Anima Anandkumar.
\newblock Fourier neural operator with learned deformations for pdes on general geometries.
\newblock {\em Journal of Machine Learning Research}, 24(388):1--26, 2023.

\bibitem{zhu2023reliable}
Min Zhu, Handi Zhang, Anran Jiao, George~Em Karniadakis, and Lu~Lu.
\newblock Reliable extrapolation of deep neural operators informed by physics or sparse observations.
\newblock {\em Computer Methods in Applied Mechanics and Engineering}, 412:116064, 2023.

\bibitem{zou2024neuraluq}
Zongren Zou, Xuhui Meng, Apostolos~F Psaros, and George~E Karniadakis.
\newblock {NeuralUQ}: A comprehensive library for uncertainty quantification in neural differential equations and operators.
\newblock {\em SIAM Review}, 66(1):161--190, 2024.

\bibitem{kovachki2021universal}
Nikola Kovachki, Samuel Lanthaler, and Siddhartha Mishra.
\newblock On universal approximation and error bounds for {Fourier} neural operators.
\newblock {\em Journal of Machine Learning Research}, 22(290):1--76, 2021.

\bibitem{wen2022u}
Gege Wen, Zongyi Li, Kamyar Azizzadenesheli, Anima Anandkumar, and Sally~M Benson.
\newblock {U-FNO—An} enhanced fourier neural operator-based deep-learning model for multiphase flow.
\newblock {\em Advances in Water Resources}, 163:104180, 2022.

\bibitem{lu2022multifidelity}
Lu~Lu, Rapha{\"e}l Pestourie, Steven~G Johnson, and Giuseppe Romano.
\newblock Multifidelity deep neural operators for efficient learning of partial differential equations with application to fast inverse design of nanoscale heat transport.
\newblock {\em Physical Review Research}, 4(2):023210, 2022.

\bibitem{kopanivcakova2024deeponet}
Alena Kopani{\v{c}}{\'a}kov{\'a} and George~Em Karniadakis.
\newblock Deepo{N}et based preconditioning strategies for solving parametric linear systems of equations.
\newblock {\em SIAM Journal on Scientific Computing}, 47(1):C151--C181, 2025.

\bibitem{zhang2022hybrid}
Enrui Zhang, Adar Kahana, Alena Kopani{\v{c}}{\'a}kov{\'a}, Eli Turkel, Rishikesh Ranade, Jay Pathak, and George~Em Karniadakis.
\newblock Blending neural operators and relaxation methods in {PDE} numerical solvers.
\newblock {\em Nature Machine Intelligence}, pages 1--11, 2024.

\bibitem{rahaman2019spectral}
Nasim Rahaman, Aristide Baratin, Devansh Arpit, Felix Draxler, Min Lin, Fred Hamprecht, Yoshua Bengio, and Aaron Courville.
\newblock On the spectral bias of neural networks.
\newblock In {\em International conference on machine learning}, pages 5301--5310. PMLR, 2019.

\bibitem{lee2024automatic}
Youngkyu Lee, Shanqing Liu, Jerome Darbon, and George~Em Karniadakis.
\newblock Automatic discovery of optimal meta-solvers via multi-objective optimization.
\newblock {\em arXiv preprint arXiv:2412.00063}, 2024.

\bibitem{bayliss1980radiation}
Alvin Bayliss and Eli Turkel.
\newblock Radiation boundary conditions for wave-like equations.
\newblock {\em Communications on Pure and applied Mathematics}, 33(6):707--725, 1980.

\bibitem{turkel1998absorbing}
Eli Turkel and Amir Yefet.
\newblock Absorbing pml boundary layers for wave-like equations.
\newblock {\em Applied Numerical Mathematics}, 27(4):533--557, 1998.

\bibitem{turkel2013compact}
Eli Turkel, Dan Gordon, Rachel Gordon, and Semyon Tsynkov.
\newblock Compact 2d and 3d sixth order schemes for the helmholtz equation with variable wave number.
\newblock {\em Journal of Computational Physics}, 232(1):272--287, 2013.

\bibitem{ogilvy1987wave}
J.A. Ogilvy.
\newblock Wave scattering from rough surfaces.
\newblock {\em Reports on Progress in Physics}, 50(12):1553, 1987.

\bibitem{voronovich2013wave}
Alexander~G Voronovich.
\newblock {\em Wave scattering from rough surfaces}, volume~17.
\newblock Springer Science \& Business Media, 2013.

\bibitem{kahana2023geometry}
Adar Kahana, Enrui Zhang, Somdatta Goswami, George Karniadakis, Rishikesh Ranade, and Jay Pathak.
\newblock On the geometry transferability of the hybrid iterative numerical solver for differential equations.
\newblock {\em Computational Mechanics}, 72(3):471--484, 2023.

\bibitem{sommerfeld1949partial}
Arnold Sommerfeld.
\newblock {\em Partial differential equations in physics}.
\newblock Academic Press, London, 1949.

\bibitem{zarmi2013general}
Asaf Zarmi and Eli Turkel.
\newblock A general approach for high order absorbing boundary conditions for the {Helmholtz} equation.
\newblock {\em Journal of Computational Physics}, 242:387--404, 2013.

\bibitem{BGT1982}
Alvin Bayliss, Max Gunzburger, and Eli Turkel.
\newblock Boundary conditions for the numerical solution of elliptic equations in exterior regions.
\newblock {\em SIAM Journal on Applied Mathematics}, 42(2):430--451, 1982.

\bibitem{gan1993finite}
H~Gan, PL~Levin, and Reinhold Ludwig.
\newblock Finite element formulation of acoustic scattering phenomena with absorbing boundary condition in the frequency domain.
\newblock {\em The Journal of the Acoustical Society of America}, 94(3):1651--1662, 1993.

\bibitem{xu1992iterative}
Jinchao Xu.
\newblock Iterative methods by space decomposition and subspace correction.
\newblock {\em SIAM review}, 34(4):581--613, 1992.

\bibitem{BarattaEtal2023}
Igor~A. Baratta, Joseph~P. Dean, J{\o}rgen~S. Dokken, Michal Habera, Jack~S. Hale, Chris~N. Richardson, Marie~E. Rognes, Matthew~W. Scroggs, Nathan Sime, and Garth~N. Wells.
\newblock {DOLFINx}: the next generation {FEniCS} problem solving environment.
\newblock preprint, 2023.

\bibitem{osti_2205494}
S.~Balay, S.~Abhyankar, M.~F. Adams, S.~Benson, J.~Brown, P.~Brune, K.~Buschelman, E.~M. Constantinescu, L.~Dalcin, A.~Dener, V.~Eijkhout, J.~Faibussowitsch, W.~D. Gropp, V.~Hapla, T.~Isaac, P.~Jolivet, D.~Karpeev, D.~Kaushik, M.~G. Knepley, F.~Kong, S.~Kruger, D.~A. May, L.~Curfman McInnes, R.~Tran Mills, L.~Mitchell, T.~Munson, J.~E. Roman, K.~Rupp, P.~Sanan, J.~Sarich, B.~F. Smith, S.~Zampini, H.~Zhang, and J.~Zhang.
\newblock {PETSc/TAO} users manual (rev. 3.20).
\newblock 11 2023.

\bibitem{paszke2019pytorch}
Adam Paszke, Sam Gross, Francisco Massa, Adam Lerer, James Bradbury, Gregory Chanan, Trevor Killeen, Zeming Lin, Natalia Gimelshein, Luca Antiga, et~al.
\newblock {PyTorch}: An imperative style, high-performance deep learning library.
\newblock {\em Advances in Neural Information Processing Systems}, 32, 2019.

\bibitem{kingma2014adam}
Diederik~P Kingma and Jimmy Ba.
\newblock Adam: A method for stochastic optimization.
\newblock {\em arXiv preprint arXiv:1412.6980}, 2014.

\end{thebibliography}

% appendix
\newpage
\appendix

\section{Experiments of 2D scattering problems}
\label{sec:app2d}
\begin{figure}
     \centering
     \begin{subfigure}[b]{0.65\textwidth}
         \centering
         \includegraphics[width=\textwidth]{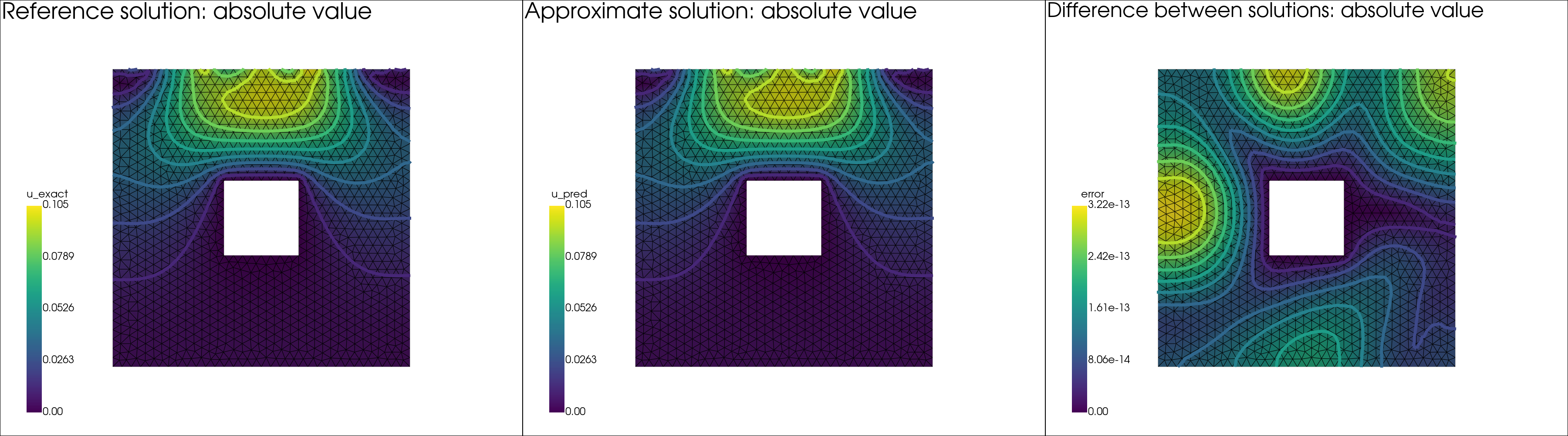}
         \caption{Reference solution / Approximate solution / Absolute error~(Square scatterer)}
     \end{subfigure}
     \begin{subfigure}[b]{0.3\textwidth}
         \centering
         \includegraphics[width=\textwidth]{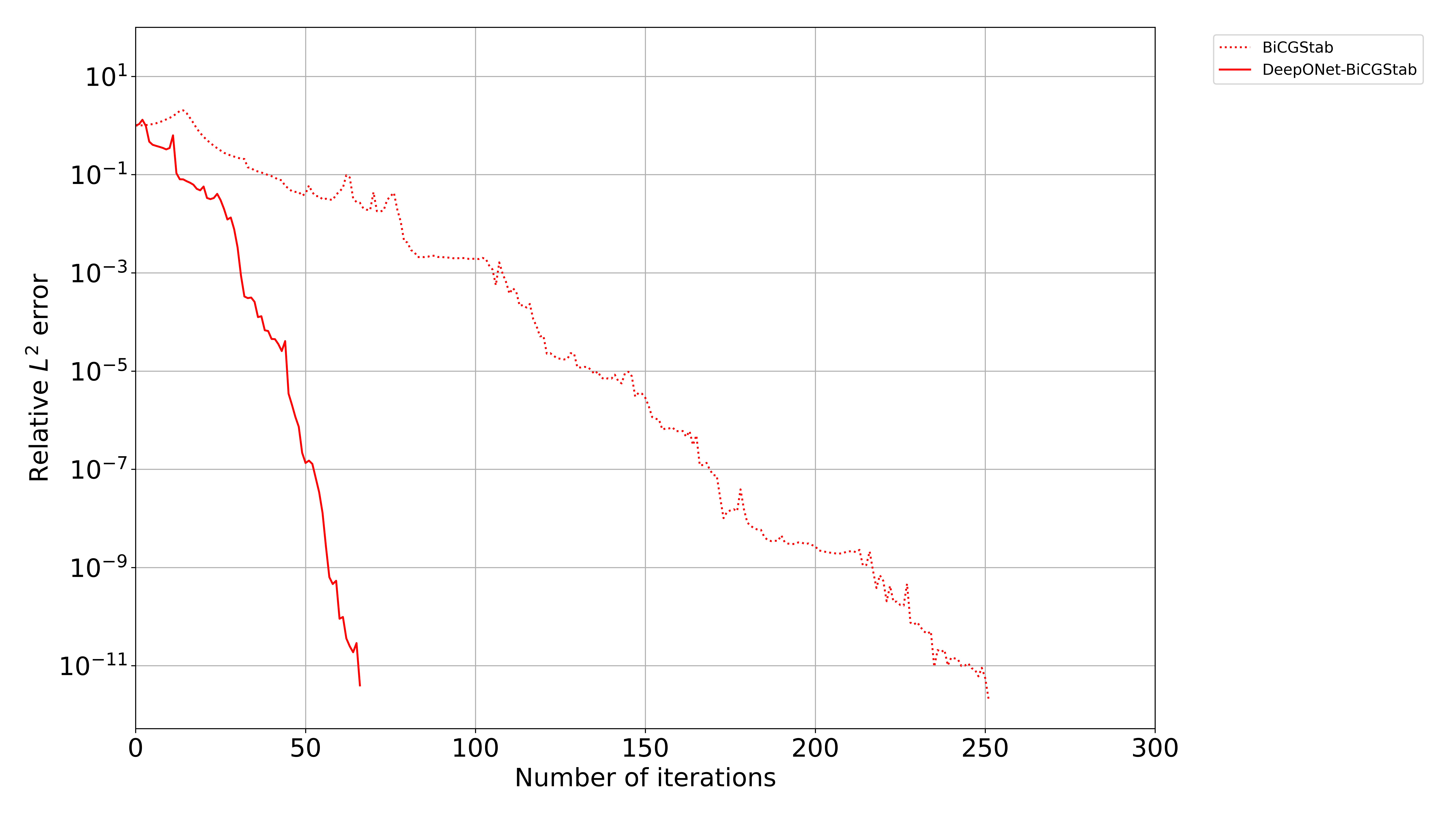}
         \caption{Relative $L^{2}$ error w.r.t. the number of iterations}
     \end{subfigure}
     \begin{subfigure}[b]{0.65\textwidth}
         \centering
         \includegraphics[width=\textwidth]{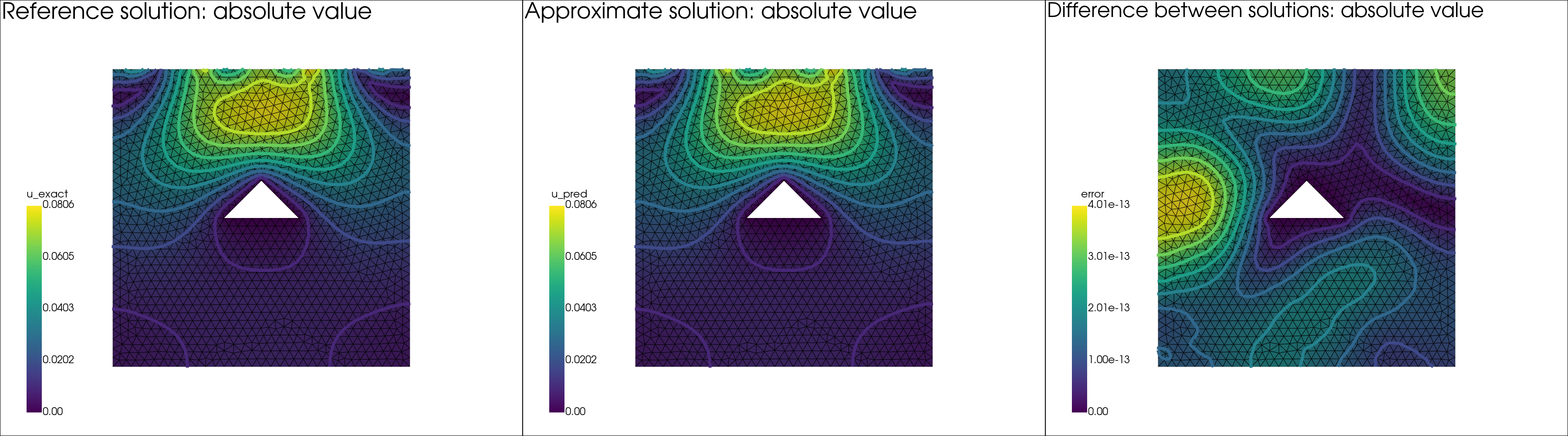}
         \caption{Reference solution / Approximate solution / Absolute error~(Triangle scatterer)}
     \end{subfigure}
     \begin{subfigure}[b]{0.3\textwidth}
         \centering
         \includegraphics[width=\textwidth]{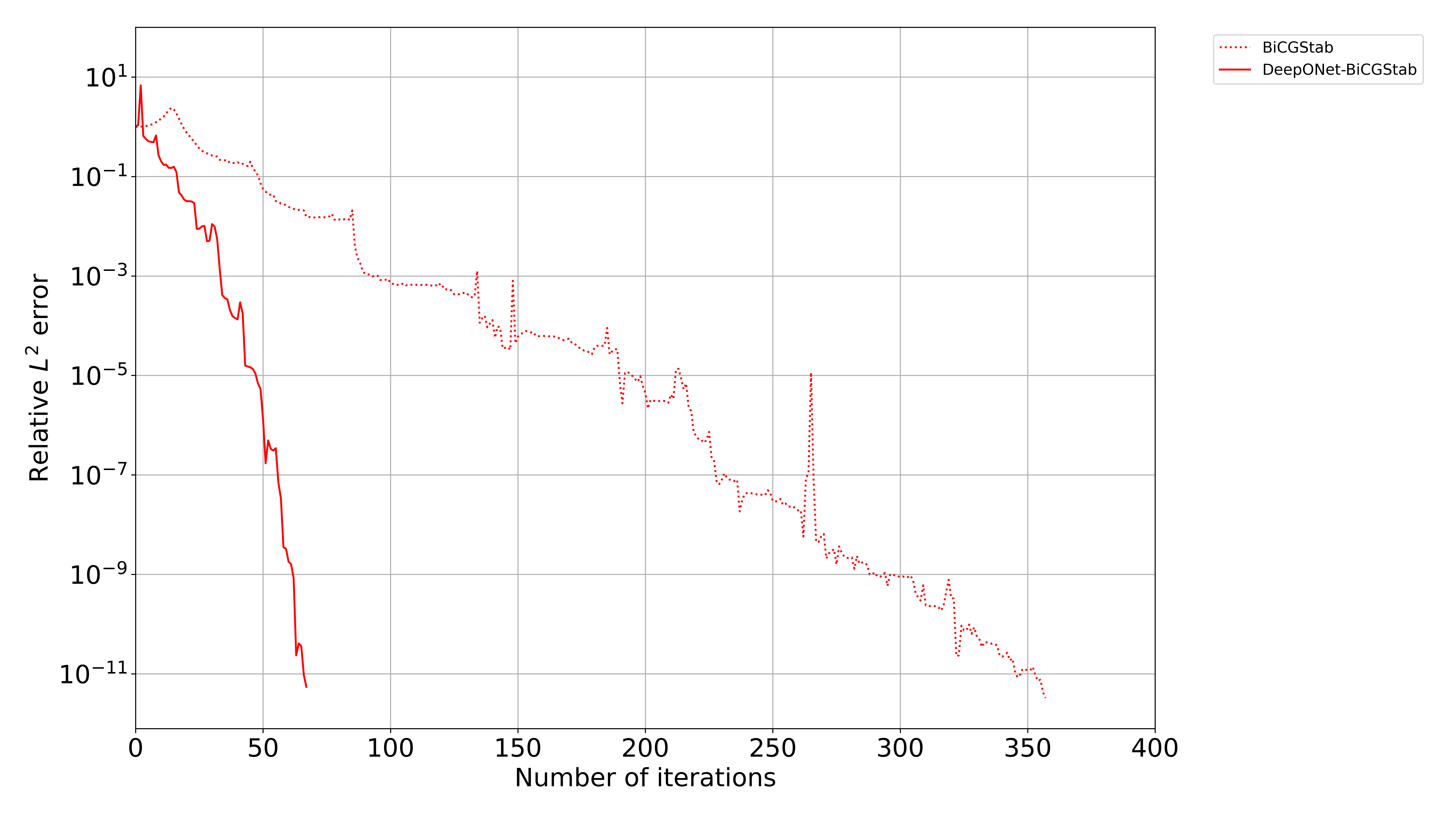}
         \caption{Relative $L^{2}$ error w.r.t. the number of iterations}
     \end{subfigure}
     \caption{Numerical results for the standard BiCGSTAB~(dotted line) and the DeepONet-BiCGSTAB~(solid line) in solving the 2D scattering problem on $[0, 1]^{2}$ with wave number $k=6$ and square scatterer~(in (a) and (b)) and triangle scatterer~(in (c) and (d)) placed at the center of the domain, respectively. 
     Note that the NO employed in HINTS is trained on non-scattering problems and approximates the solution operator of Eq. \eqref{eq:problem} with $\Omega_{scat} = \emptyset$. The ratio between the relaxation and neural operator is $3$ and the number of trunk-basis is $p=32$.}
     \label{fig:2d}
\end{figure}

% \begin{figure}
%      \centering
%      \begin{subfigure}[b]{0.22\textwidth}
%          \centering
%          \includegraphics[width=\textwidth]{figs/2D/case1_k.png}
%          \caption{Wave number $k(x, y)$.}
%      \end{subfigure}
%      \begin{subfigure}[b]{0.7\textwidth}
%          \centering
%          \includegraphics[width=\textwidth]{figs/2D/case1_u_ref.png}
%          \caption{Reference solution.}
%      \end{subfigure}
%      \begin{subfigure}[b]{1\textwidth}
%          \centering
%          \includegraphics[width=\textwidth]{figs/2D/case1_jacobi.png}
%          \caption{Using Jacobi.}
%      \end{subfigure}
%      \begin{subfigure}[b]{1\textwidth}
%          \centering
%          \includegraphics[width=\textwidth]{figs/2D/case1_hints.png}
%          \caption{Using HINTS.}
%      \end{subfigure}
%      \caption{Comparison between using the standard Jacobi method (in (c)) and using the HINTS (in (d)) in solving the 2D scattering problem on $[0, 1]^2$ with a square scatterer (of size $0.125\times 0.125$) placed at the center of the domain. 
%      % (a) shows the wave number $k(x, y)$ and (b) shows the reference solution. 
%      % The Jacobi solution diverges, as expected, for a non-positive-definite problem, while the solution yielded by HINTS is accurate to machine precision. 
%      Note that the NO employed in HINTS is trained on non-scattering problems and approximates the solution operator of Eq. \eqref{eq:problem} with $\Omega_{scat} = \emptyset$.}
%      \label{fig:2d_case1}
% \end{figure}

We solve the 2D scattering problem on a square domain $\Omega := [0, 1]^2$.
The approximate Sommerfeld boundary condition, $\frac{\partial u}{\partial \text{n}} +i k u= 0$, is imposed on the left, right and bottom boundaries, while a Neumann boundary condition is imposed on the top boundary, i.e. $\frac{\partial u}{\partial \text{n}}(x) = g(x), x\in \partial\Omega$, and a Dirichlet boundary condition is imposed on the scatterer, denoted by $\Omega_{scat}$.
Unless stated otherwise, we discretize the domain uniformly with a $33 \times 33$ mesh, set the incoming wave $g(x)=\sin(3\pi x), x\in[0, 1]$ and the forcing term $f(x, y) = 0, x, y\in[0, 1]$.  We randomly sample the wave number $k(x, y)$ from a 2D Gaussian random field with mean $6.0$ and the radial basis function~(RBF) kernel function:
\begin{equation*}
\label{eq:kernel}
    K(\mathbf{x}, \mathbf{x}^\prime) = s^2\exp(-\frac{||\mathbf{x}-\mathbf{x}^\prime||^2}{2l^2}), \mathbf{x}, \mathbf{x}^\prime \in \Omega,
\end{equation*}
where $s^{2}$ is the variance and is set to $0.5$, $l$ denotes the correlation length and is set to $0.3$, and $||\cdot||$ denotes the $\ell_2$ vector norm.
We stress that this approach applies to any discretization and incoming wave function.

We first consider a square scatterer of size $0.25 \times 0.25$ centered at $(x, y) = (0.5, 0.5)$. 
The NO approximates the solution operator of~\eqref{eq:problem}, which maps the wave number $k$, the (non-zero) complex forcing term $f$, and the incoming wave $g$ to its sought solution $u$.
However, the NO is trained to learn the solution operator of the 2D Helmholtz equation without the scatterer in which $\Omega_{scat} = \emptyset$, and used in HINTS to solve the scattering problem in which $\Omega_{scat} \neq \emptyset$. 
Specifically, the training data of the NO are obtained from solving~\eqref{eq:problem} while ignoring the scatterer and its associated Dirichlet boundary condition using randomly generated samples of $k$, $f$, and $g$.
The details of the data generation, the architecture of DeepONets, and the training are given in~\ref{sec:details}.
The success of the hybrid preconditioner in solving the scattering problem with a NO trained to solve non-scattering ones demonstrates the high flexibility and the extrapolation capability of our approach.
The comparison between the standard BiCGStab method and DeepONet-BiCGStab is presented~\Cref{fig:2d}(a), (b).

We next consider a scatterer with a different geometry. Specifically, we solve the scattering problem with a triangular scatterer placed at the center of the domain and defined by the coordinates of its three vertices: $(0.375, 0.5)$, $(0.625, 0.5)$, and $(0.5, 0.625)$. 
We test the hybrid preconditioning strategy with the same wave number $k(x, y)$ and incoming wave $g(x)$.
The NO is chosen as the one trained on non-scattering problems as in previous cases. 
The comparison between the standard BiCGStab method and DeepONet-BiCGStab is presented~\Cref{fig:2d}(c), (d).
We observe a similar divergence of BiCGStab and convergence of DeepONet-BiCGStab as in the square scatterer case.

% \subsection{A different mesh}
% \begin{figure}
%      \centering
%      \begin{minipage}{.22\linewidth}
%     \centering
%     \subcaptionbox{Reference solution}
%       {\includegraphics[width=\textwidth]{figs/2D/case5_u_ref.png}}
%     \subcaptionbox{The wave number}
%       {\includegraphics[width=\textwidth]{figs/2D/case5_k.png}}
%   \end{minipage}
%      \begin{minipage}{.66\linewidth}
%     \centering
%     \subcaptionbox{Using Jacobi}
%       {\includegraphics[width=\textwidth]{figs/2D/case5_jacobi.png}}

%     \subcaptionbox{Using HINTS}
%       {\includegraphics[width=\textwidth]{figs/2D/case5_hints.png}}
%   \end{minipage}
%      \caption{Using HINTS to solve the scattering problem discretized on a $65\times 65$ uniform mesh, while the NO employed in HINTS is trained on a $33\times 33$ uniform mesh.}
%      \label{fig:2d_case5}
% \end{figure}

\section{Details of data generation and the training of neural operators (NOs)}
\label{sec:details}

In this section, we present additional details of the numerical examples, including data generation, the architecture of DeepONets, and their training. In this work, the Adam optimizer \cite{kingma2014adam} with a constant learning rate $1\times10^{-4}$ is used to train DeepONets, and the PyTorch module \cite{paszke2019pytorch} is used for machine learning related computation.

\subsection{2D scattering problems}

In~\ref{sec:app2d}, the 2D Gaussian random field (GRF) from which samples of the wave number $k(x, y)$ are drawn has mean six and radial basis function~(RBF) kernel function with $l=0.3$ and $s=0.5$. 
The 1D GRF for the incoming wave $g(x)$ has mean zero and RBF kernel function with $l=0.1$ and $s=1$. 
For the forcing term $f(x, y)$, since it is a complex-valued function, samples of its real part and samples of its imaginary part are drawn independently from the same 2D GRF, which has mean zero and RBF kernel function with $l=0.1$ and $s=1$. 
For the training of NOs, we take $20,000$ samples of $k$, $f$, and $g$, and solve the 2D complex Helmholtz equation to obtain the data of $u$.
For the NO on non-scattering problems, the data of $u$ are obtained by solving~\eqref{eq:problem} with $\Omega_{scat} = \emptyset$, while for the NO on scattering problems, the data of $u$ are obtained by solving~\eqref{eq:problem} with $\Omega_{scat}=[0.4375, 0.5625]^2$ and associated Dirichlet boundary condition.
We note that the forcing term is not assumed to be zero in the training of NOs because the NO approximates the solution operator of the complex Helmholtz equation. However, for the physical scattering problem described by the Helmholtz equation, the forcing term is set equal to zero.

In this work, the DeepONet employed in HINTS takes the wave number and the forcing term as input.
In~\ref{sec:app2d}, the branch net of DeepONet is a 2D convolutional neural network (CNN) (input dimension $33\times 33$, channels number $[3, 40, 60, 100, 180$], kernel size $3\times 3$, stride $2\times 2$, valid padding, ReLU activation function) followed by a fully-connected neural network (FNN) (width $[180, 256, 256, 256]$, ReLU activation function for all hidden layers).
Its trunk net is a FNN (width $[2, 256, 256, 128]$, Leaky ReLU activation function).
The output of the branch net is divided into two groups, and the real and imaginary parts of the output of the DeepONet is constructed by taking the inner product with the output of the trunk net.
The input to the branch is the concatenation of the values of the wave number and the real and imaginary parts of the forcing term on a mesh.
NOs in~\ref{sec:app2d} are trained for $10,000$ epochs with batch size $500$.
DeepONet was trained until the relative $L^{2}$ error was lower than $8\%$ or the training loss plateaued around $100$ epochs, which took $1629.8$ seconds.

\subsection{3D scattering problems}

In~\Cref{sec:4}, the 3D Gaussian random field (GRF) from which samples of the wave number $k(x, y, z)$ are drawn has mean six and RBF kernel function with $l=0.3$ and $s=0.2$. Similarly, we reject samples of $k$ whose minimal values are smaller than or equal to three to ensure that the wave number is not too small.
The 2D GRF for the incoming wave $g(x, y)$ has mean zero and RBF kernel function with $l=0.1$ and $s=1$. 
The 3D GRFs for the real and imaginary parts of the forcing term $f(x, y, z)$ have mean zero and RBF kernel function with $l=0.1$ and $s=1$. 
For the training of NOs, we take $30,000$ samples of $k$, $f$, and $g$, and solve the 3D complex Helmholtz equation without the scatterer~($\Omega_{scat} = \emptyset$ in~\eqref{eq:problem}) to obtain the data of $u$.

For the DeepONet which takes as input the wave number and the forcing term, its branch net is a 3D CNN (input dimension $17 \times 17 \times 17$, channels number $[3, 40, 40, 60]$, kernel size $3\times 3$, stride $2\times 2$, same padding in the second convolutional layer and valid padding in the rest, ReLU activation function) followed by a FNN (width $[1620, 256, 256, 256]$, ReLU activation function for all hidden layers).
Its trunk net is an FNN (width $[3, 256, 256, 128]$, Leaky ReLU activation function for all hidden layers).
The DeepONet's output is generated the same way as in the 2D case.
The input to the branch is the concatenation of the values of the wave number and the real and imaginary parts of the forcing term on a mesh.
Note that The DeepONet is trained for $20,000$ epochs with batch size $500$. DeepONet was trained until the relative $L^{2}$ error was lower than $8\%$ or the training loss plateaued around $100$ epochs, which took  $55817.3$ seconds.

\section{Trunk basis (TB) hybridization approach}
\label{sec:tb_approach}
Let us assume that the pre-trained DeepONet consists of $p$ number of basis functions.
Given a set of coordinates $\{\mathbf{x}_{i}\}_{i=1}^{n}$, 
we can compute a matrix $\mathbf{T} \in \mathbb{R}^{n \times p}$ using the trunk network, i.e. the $(i,j)$-th component of $\mathbf{T}$ is given by
\begin{equation}\label{eq:prolongation}
    [\mathbf{T}]_{ij} = T_{j}(\mathbf{x}_{i}),
\end{equation}
where $T_{j}(\mathbf{x}_{i})$ denotes the $j$-th component of the output of the trunk network evaluated at $\mathbf{x}_{i}$.
Using an appropriate selection algorithm, we choose $k \ll p$ number of columns of $\mathbf{T}$ and construct the prolongation operator $\mathbf{P} \in \mathbb{R}^{n \times k}$.
Natural-order selection, random selection, and selection from the smallest singular value are possible algorithms.
In this work, we employed natural-order selection.
The restriction operator $\mathbf{R}$ is naturally defined as the adjoint operator of $\mathbf{P}$, i.e., $\mathbf{R}:=\mathbf{P}^{\ast}$.
Using $\mathbf{R}$ and $\mathbf{P}$, the second preconditioner $\mathbf{M}_{2}$ for Krylov methods introduced in~\eqref{eq:hybrid} is constructed as
\begin{equation*}
    \mathbf{M}_{2} := \mathbf{P}(\mathbf{R}\mathbf{A}\mathbf{P})^{-1}\mathbf{R} = \mathbf{P}\mathbf{A}_{c}^{-1}\mathbf{R}.
\end{equation*}
Since the new preconditioner $\mathbf{M}_{2}$ is linear, the preconditioned iteration~\eqref{eq:hybrid} can be applied to a Krylov method.
In addition, to achieve a well-conditioned operator $\mathbf{A}_{c}$, it is crucial to ensure that the operator $\mathbf{P}$ has full rank and orthogonal columns.
One promising approach is to perform the QR decomposition of $\mathbf{P}$, i.e. $\mathbf{P}=\mathbf{Q}\mathbf{R}$, and use $\mathbf{Q}$ as the prolongation operator instead of $\mathbf{P}$.
A sketch of the construction of Krylov-based meta-solvers is presented in~\Cref{fig:krylov}. 

For more details, we also refer to~\cite{zhang2022hybrid}, which is the first work on hybridizing DeepONet with relaxation methods, and~\cite{kopanivcakova2024deeponet}, which is the first work of hybridizing DeepONet with Krylov methods.

\section{No training is required for non-homogeneous boundary conditions}
\label{sec:homogenuous}
The hybrid preconditioner defined in~\Cref{sec:hints} acts on the residual $\mathbf{r}^{(i)}$ at each iteration, not the right-hand side $\mathbf{f}$.
Let us assume that we are solving the following boundary value problem:
\begin{equation}
\label{eq:poisson}
\left\{
\begin{aligned}
 - u''(x) &=f \text{ in } \Omega = (0,1), \\
u(0) &= g(0), \\
\frac{\partial u}{\partial n}(1) &= T(1).
\end{aligned}
\right.
\end{equation}
Under a finite element (or finite difference) discretization, the domain $\Omega$ is uniformly discretized as 
\begin{equation*}
0=x_{0}< x_{1}< x_{2} < x_{3} < x_{4}=1,
\end{equation*}
and the problem~\eqref{eq:poisson} is formulated as
\begin{align}
\label{eq:fem}
\frac{1}{h^{2}}\begin{bmatrix}
1&0&0&0&0\\
-1&2&-1&0&0\\
0&-1&2&-1&0\\
0&0&-1&2&-1\\
0&0&0&-1&1\\
\end{bmatrix}
\begin{bmatrix}
u_{0}\\
u_{1}\\
u_{2}\\
u_{3}\\
u_{4}\\
\end{bmatrix}
&=
\begin{bmatrix}
g(x_0)\\
f(x_1)\\
f(x_2)\\
f(x_3)\\
f(x_4)-T(x_4)\\
\end{bmatrix},\\
\Rightarrow \mathbf{A} \mathbf{u} &= \mathbf{f}, \nonumber
\end{align}
where $h$ is the given mesh size.
From the formulation~\eqref{eq:fem}, the boundary values~$(u_{0}, u_{4})$ are directly determined by
\begin{equation*}
\left\{
\begin{aligned}
u_0 &= g(x_0),\\
u_4 &= f(x_4) - T(x_4) +u_3.
\end{aligned}
\right.
\end{equation*}
When we utilize any iterative solver, the $i$-th residual $\mathbf{r}^{(i)}$ is computed as follows:
\begin{equation*}
\label{eq:residual}
\mathbf{r}^{(i)} = \mathbf{f} - A\mathbf{u}^{(i)} = 
\begin{bmatrix}
g(x_0) - u_0^{(i)} = 0\\
f(x_{1}) - 1/h^{2} (-u_{0}^{(i)} + 2 u_{1}^{(i)} - u_{2}^{(i)})\\
f(x_{2}) - 1/h^{2} (-u_{1}^{(i)} + 2 u_{2}^{(i)} - u_{3}^{(i)})\\
f(x_{3}) - 1/h^{2} (-u_{2}^{(i)} + 2 u_{3}^{(i)} - u_{4}^{(i)})\\
f(x_4)-T(x_4) +u_{3}^{(i)} - u_{4}^{(i)} = 0\\
\end{bmatrix}.
\end{equation*}
Since the $i$-th residual $\mathbf{r}^{(i)}$ has a value of $0$ on the boundary, independent of $g$ and $T$, it is enough for DeepONet to be trained on homogenous problems.
% Since the $i$-th residual $\mathbf{r}^{(i)}$ has a value of $0$ on the boundary, independent of $g$ and $T$, it is enough for DeepONet to be trained on homogenous problems as a foundation model.
This fact avoids the need to generate the incoming wave $g$ to create training data for scattering problem~\eqref{eq:problem}, reducing the memory required and simplifying the training of DeepONet.

\end{document}